\documentclass[accepted]{uai2026}

\usepackage[american]{babel}
\usepackage{natbib} 
\usepackage{mathtools} 
\usepackage{booktabs} 
\usepackage{tikz} 
\usepackage{algorithm}
\usepackage{algorithmic}
\usepackage{amsmath}
\usepackage{amsthm}
\usepackage[switch]{lineno}
\usepackage{amssymb}
\usepackage{subfig}

\title{Revisiting TD Target Aggregation under Uncertainty in Q-Learning}

\author[1]{\href{mailto:lz23b@fsu.edu}{Lipeng~Zu}{}}
\author[1]{\href{mailto:xzhang14@fsu.edu}{Xiaonan~Zhang}{}}

\affil[1]{%
    Computer Science Department\\
    Florida State University\\
    Tallahassee, Florida, USA
}
  
\begin{document}
\maketitle

\begin{abstract}
Deep Q-Networks (DQNs) learn value functions through bootstrapped temporal-difference updates, where future returns are approximated using a greedy maximization over next-state action values. While effective, this aggregation rule is inherently sensitive to estimation noise: when Q-values are uncertain, the maximization operator deterministically favors the largest estimate, regardless of its reliability, leading to amplified errors through bootstrapping. In this work, we propose the \textbf{S}uccessor Rollout \textbf{A}ggregation \textbf{D}eep \textbf{Q}-Network (SADQ), a simple modification to Q-learning that regularizes how the TD target is formed. SADQ uses one-step rollout predictions from a learned dynamics model to guide the comparison among candidate next-state actions, introducing additional structure into the aggregation step without altering the underlying learning framework. The resulting mixed Bellman update attenuates unreliable maxima while preserving the standard fixed point under diminishing model error. We provide theoretical analysis showing that SADQ reduces bootstrap-induced overestimation in a pointwise manner. Empirically, SADQ consistently improves training stability across classical control tasks, real-world vector-based environments, and Atari benchmarks when compared to strong DQN variants.
\end{abstract}

\section{Introduction}\label{sec:intro}
Q-learning and its deep variants, such as Deep Q-Networks (DQN), learn action-value functions through bootstrapped temporal-difference (TD) updates~\citep{mnih2015human, liang2022reducing}. In these methods, future returns are approximated using a greedy backup, $\max_{a'} Q(s', a')$, rather than being directly observed~\citep{moore1993prioritized, ghiassian2020gradient}. This mechanism has proven effective in practice, but it also introduces a well-known source of instability when combined with function approximation~\citep{maei2009convergent, lee2019target, rowland2023statistical}.

A large body of work has focused on improving the quality of Q-value estimation. Strategies such as prioritized sampling~\citep{schaul2015prioritized}, multi-step returns~\citep{mnih2016asynchronous}, architectural refinements including Double and Dueling DQN~\citep{van2016deep, wang2016dueling}, and distributional methods~\citep{bellemare2017distributional, dabney2018distributional} have significantly improved empirical performance and stability. These methods primarily aim to reduce estimation uncertainty or better characterize predictive dispersion at the level of individual value predictions. However, even when value estimates are improved, the TD target is still formed by a hard maximization over next-state actions. This maximization implicitly assumes that the relative ordering of Q-values is reliable. In practice, especially during learning, Q-values are often noisy and uncertain. Under such conditions, selecting actions solely based on their maximum estimated value can be brittle: the greedy operator may favor actions that appear optimal due to transient overestimation rather than genuine long-term advantage. This uncertainty is recursively amplified through bootstrapped updates, leading to biased targets and unstable learning dynamics. 

This observation suggests that the challenge is not only about estimating Q-values accurately, but also about how candidate actions are ranked under uncertainty when forming the TD target. In particular, the standard greedy backup lacks any mechanism to assess the reliability of competing action values. A natural question then arises: can we introduce additional structure into the aggregation step of the TD update, without abandoning the simplicity and efficiency of Q-learning? One source of such structure comes from predicting the immediate consequences of actions under the environment dynamics. Recent advances in learned dynamics models~\citep{ha2018recurrent, hafner2025mastering} have shown that short-horizon rollouts can provide meaningful information about action outcomes. Yet, in most existing work, these models are used for planning or data generation, rather than for shaping the TD target itself. 

In this paper, we propose \emph{Successor Rollout Aggregation Deep Q-Networks} (SADQ), a simple modification to the TD target construction that incorporates one-step rollout information into action comparison. Instead of relying solely on greedy value maximization, SADQ uses rollout-based evaluations to guide the selection of the next-state action used in the target. Importantly, the rollout is not used to replace Q-values, but to regularize the aggregation process, reducing sensitivity to unreliable maxima while preserving the standard bootstrap framework.

Our contributions can be summarized as follows:
\begin{itemize}
    \item We identify brittle action aggregation under noisy value estimates as a structural source of instability in Q-learning.
    \item We propose SADQ, a rollout-guided aggregation mechanism that regularizes greedy TD target construction without altering the underlying learning paradigm.
    \item We provide theoretical analysis showing that the resulting update reduces maximization-induced bias while preserving the optimal fixed point under diminishing model error.
    \item We demonstrate consistent empirical improvements across a range of benchmark and real-world decision-making tasks.
\end{itemize}

\section{Related Work}
\paragraph{Stabilization Strategies for Q Estimation.} 
The core instability in Q-learning arises from the bootstrapped target, where the hard maximization operator introduces systematic overestimation bias and amplifies estimation variance~\citep{mnih2015human, sutton2018reinforcement}. Early work such as Double DQN mitigates overestimation bias in Q-value predictions~\citep{van2016deep}, while Dueling DQN separates the state-value and advantage functions~\citep{wang2016dueling}. Advancing further, distributional RL techniques, such as C51~\citep{bellemare2017distributional} and QR-DQN~\citep{dabney2018distributional}, model the full return distribution before applying maximization, thereby altering how uncertainty propagates through the max operator~\citep{tang2022nature, schwarzer2023bigger, kastner2025categorical}. Bayesian approaches also emerge, integrating prior knowledge and adaptive exploration strategies into the DQN framework~\citep{cao2012bayesian}. Complementary to these innovations, sampling techniques are used to enhance data efficiency and exploration at the sample level~\citep{osband2015bootstrapped, schaul2015prioritized, elvira2021advances}. Despite these advances, most existing methods eventually revert to the max operation for Q updates. In contrast, our work focuses explicitly on restructuring the max-based TD target itself.

\paragraph{Model-Based and Generative Approaches in RL.}
Model-based RL exploits learned dynamics models for planning and decision-making~\citep{ha2018recurrent, schwarzer2020data, mondal2023efficient}. For example, Recurrent State-Space Model (RSSM)~\citep{hafner2019learning, hafner2025mastering} combines variational inference with recurrent dynamics to infer compact latent states. Subsequent approaches such as RePo~\citep{zhu2023repo} and Denoised MDP~\citep{wang2022denoised} remove pixel reconstruction and emphasize reward-relevant features to improve tractability and task alignment. To enhance generalization and robustness, several methods introduce information bottlenecks through mutual information regularization across latent representations~\citep{bai2021dynamic, saanum2023reinforcement, tang2023understanding}. Diffusion-based generative models have recently been incorporated into RL as powerful tools for trajectory modeling and synthetic data generation~\citep{janner2022planning, lu2023synthetic, wang2026off}. Some approaches integrate policy or value guidance into diffusion processes to bias generation toward task-relevant behaviors~\citep{rigter2024world, chen2023offline}. Others adopt energy-based objectives to guide generative modeling for RL learning~\citep{lu2023contrastive, liu2024energy}. In contrast to approaches that leverages predictive models for long-horizon planning, our method employs one-step successor rollouts and integrates them directly into TD target aggregation, thereby modifying the update rule rather than the planning process.

\paragraph{Relation to Path-Consistency Methods.}
The proposed aggregation bears conceptual similarity to path-consistency approaches such as PCZero~\citep{zhao2023generalized}, which reduce variance by enforcing consistency along selected trajectories. PCZero constructs a window over historical and search-expanded states and minimizes value variation within that window, reflecting the principle that values along an optimal path should agree~\citep{zhao2022efficient}. The mixed TD target in Eq.~(\ref{eq:mixed_target_correct}) can be viewed as a single-step analogue of this idea. Given the set of rollout-informed estimates $\{\widetilde{Q'}_\phi(s',a')\}_{a'\in\mathcal A}$, we identify the action whose predicted successor outcome is most aligned with the greedy backup, and interpolate between this action and the standard maximizer. In contrast to multi-step or window-based consistency, our approach operates at the finest temporal granularity of a single transition, directly regularizing the aggregation step in the TD update.

\paragraph{Relation to Model-based Value Expansion.}
SADQ is related to model-based value expansion methods in that both use predictive model information to improve value learning. Methods such as MVE~\citep{palenicek2022revisiting} and STEVE~\citep{buckman2018sample} use learned dynamics and reward models to construct expanded value targets by accumulating predicted rewards over rollout horizons and then bootstrapping from the terminal predicted state. In contrast, SADQ does not use the model to expand the Bellman backup over multiple rollout steps. The auxiliary model provides a one-step lookahead score for each candidate next action, which is used to refine the action-selection step inside the max-based TD target. Thus, SADQ shares the motivation of leveraging predictive structure for value learning, but differs in where the model enters the update. Value expansion modifies the target value through rollout-based reward accumulation, whereas SADQ modifies the action aggregation process while keeping the final bootstrapped value evaluation within the Q-learning framework.

\section{Preliminaries}
\subsection{Reinforcement Learning}
RL is a machine learning paradigm where an agent learns to make decisions through the interaction with the environment~\citep{matsuo2022deep, saanum2024reinforcement, zhang2024sf}. Typically, the RL problem is modeled as a Markov Decision Process (MDP), characterized  by a tuple \((S, A, P, R, \gamma)\). Here,  \(S\) represents the set of states, \(A\) indicates the set of actions, \(P\) denotes the state transition probabilities, \(R\) specifies the reward function, and \(\gamma\) serves as the discount factor. The objective of the agent is to learn an optimal policy \(\pi^*(s)\) that maximizes the expected cumulative reward over time. 

RL methods can be broadly categorized into policy-based~\citep{schulman2015trust}, value-based~\citep{mnih2015human}, and actor-critic approaches~\citep{haarnoja2018soft}. Policy-based methods  directly learn a policy \(\pi(a|s)\) that maps states to actions, whereas  value-based methods focus on estimating a value function, such as the Q-function \(Q(s, a)\), to estimate the quality of actions. Given that this work centers on DQN, our discussion primarily emphasizes value-based methods and their relevance to optimal policy learning.

\subsection{Deep Q-based RL}
As a foundational deep Q-based RL algorithm, Deep Q-Network (DQN)~\citep{mnih2015human} integrates Q-learning with deep neural networks to facilitate decision-making in high-dimensional state spaces. Particularly, DQN is comprised of two distinct neural networks: the primary Q-network and the target Q-network. The primary Q-network, simply referred to as the Q-network, approximates the Q-value $Q(s, a)$. To stabilize training, a separate target Q network, denoted as \(Q'(s,a)\), computes the target Q-value $y$ independently. The optimization objective for training the Q-network is defined as minimizing the TD error, given by: 
\begin{equation}
    \mathcal{L}_Q = \mathbb{E}_{(s,a,r,s')\in D}[y - Q(s, a)],
    \label{eq:dqn}
\end{equation}
where \((s,a,r,s')\) is the sampled transition from replay buffer \(\mathcal{D}\). The target Q-value  \( y \) in Eq.~(\ref{eq:dqn}) is computed by the target Q-network \(Q'\) and is formulated as: 
\begin{equation}
    y = r + \gamma \max_{a'} Q_{\text{target}}(s',a'),
    \label{eq:q_traget}
\end{equation}
The maximization operator in Eq.~(\ref{eq:q_traget}) determines how future value estimates are aggregated into the TD target, making it a critical component of the learning dynamics.

\section{Motivation}
\begin{figure}[t]
    \centering
    \subfloat[]{
        \includegraphics[width=0.226\textwidth]{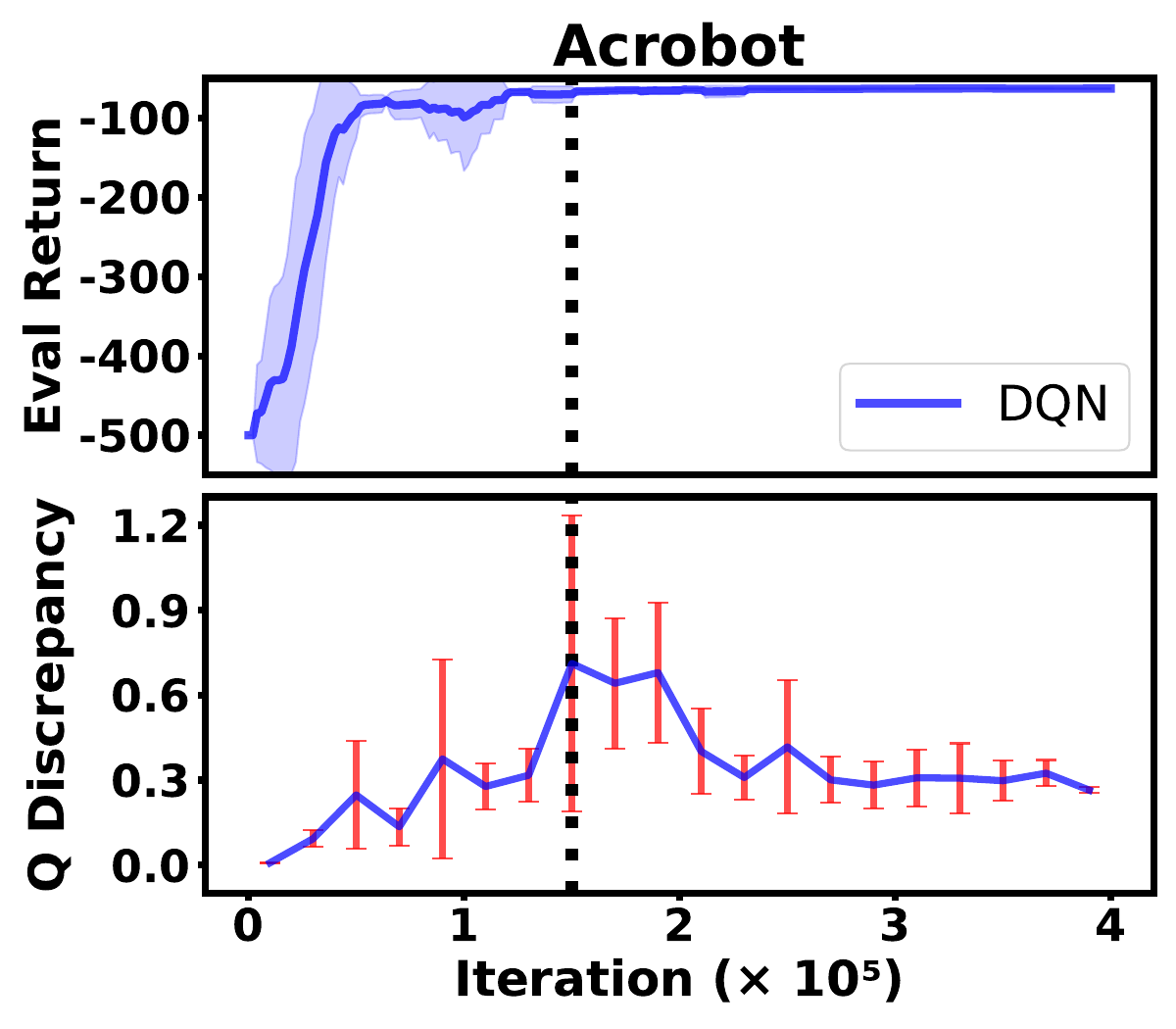} 
        \label{fig:mov_acrobot_subplot}
    }
    \hfill
    \subfloat[]{
        \includegraphics[width=0.226\textwidth]{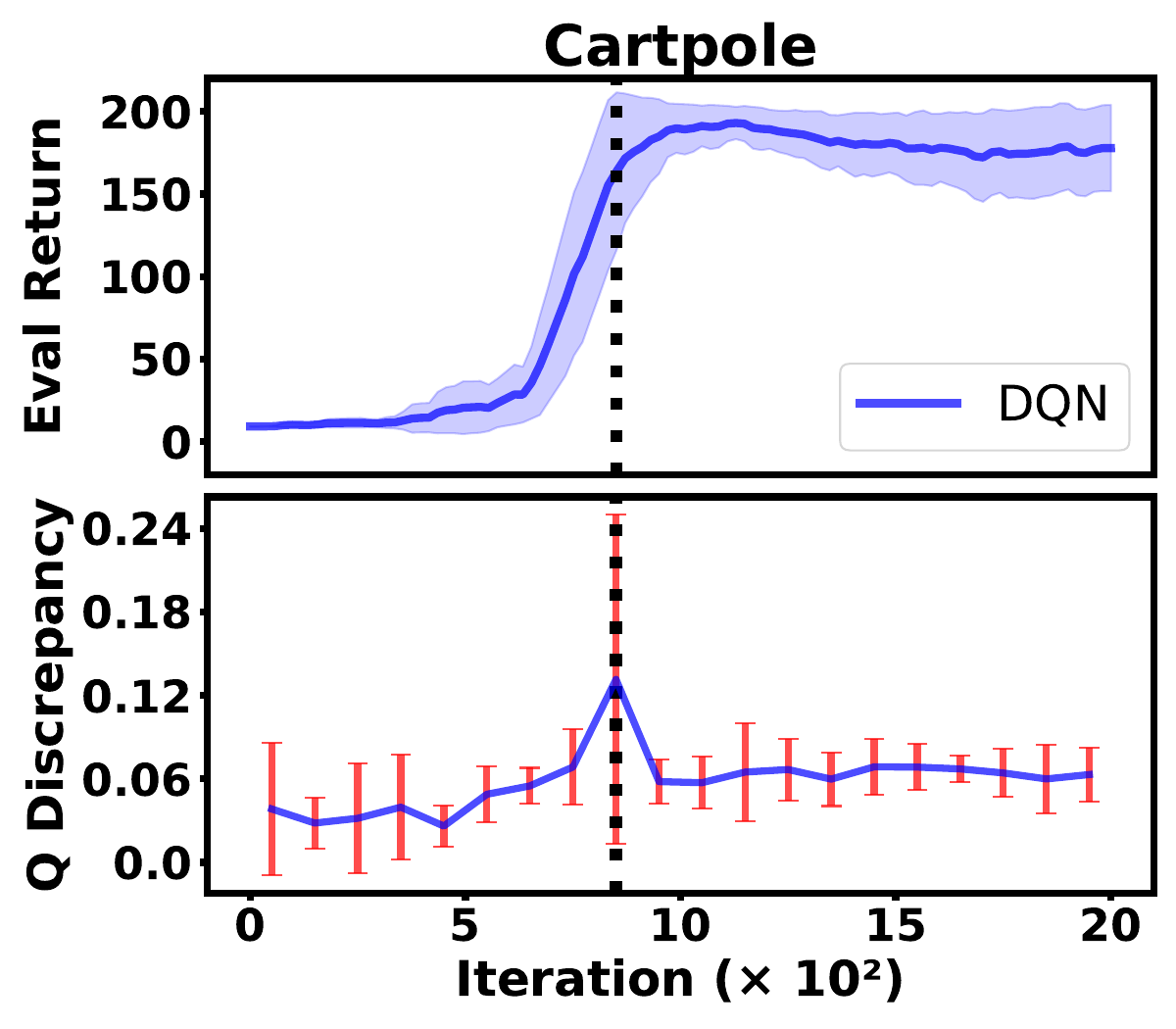}
        \label{fig:mov_cartpole_subplot}
    }
    \caption{Upper: Training performance of DQN on Acrobot and Cartpole environments. Lower: The evaluation of Q discrepancy (\(\max Q(s,a) - \min Q(s,a)\)).}
    \label{fig:mov_performance}
\end{figure}

\paragraph{\textit{Observation:}}
Fig.~\ref{fig:mov_performance} shows that DQN training typically exhibits a rapid improvement phase followed by a stable phase. To examine how value estimation evolves during this transition, we track the Q discrepancy
\(
\Delta_Q(s) = \max_a Q(s,a) - \min_a Q(s,a),
\)
which directly depends on the max-Q term used in the TD target. As shown in the lower panels, $\Delta_Q$ increases during the performance rising phase, reaches a peak near the transition point (dashed line), and then decreases to a stable level together with the evaluation return. Since $\max_a Q(s,a)$ is the quantity selected in the bootstrap target, the alignment between performance dynamics and $\Delta_Q$ indicates that training behavior is closely tied to the magnitude of the maxQ statistic. In particular, the overshoot and subsequent contraction of $\Delta_Q$ suggest that excessive amplification of the maximized Q-value is not sustained in the stable regime, motivating a closer examination of how max-Q enters the TD update.

\paragraph{\textit{Insight:}}
The key point is the large separation among action values is temporary. Once learning stabilizes, the discrepancy contracts rather than remaining large. This indicates that aggressively amplifying the largest Q-value is not a prerequisite for good performance, but instead emerges during learning when value estimates are still uncertain. During this phase, action values are noisy and continually reshaped, yet the maximization operator treats their ordering as fully reliable. By selecting a single largest estimate, the TD target can be dominated by transient peaks that are not well supported by the underlying dynamics. These peaks are then reinforced through bootstrapping, even though they eventually fade as learning progresses. The difficulty therefore lies not only in estimating Q-values accurately, but in how actions are compared when those estimates are still unreliable.

Such observation raises a simple question: can the next-state evaluation used in the TD target be made more robust than a single maximized value, without abandoning the basic structure of Q-learning? Addressing this question motivates reconsidering how predicted future consequences are incorporated into the TD update, and leads to the rollout-guided aggregation approach developed in this work.

\section{Method}
In this section, we present the Successor Rollout Aggregation Mechanism (SADQ).
We first introduce the learned dynamics model to generate one-step successor states and rewards (Section~\ref{sec:vector_model} for vector-based tasks; Section~\ref{sec:image_model} for image-based tasks). Based on these rollout predictions, we construct a structurally aggregated TD target for DQN and its distributional variants (Section~\ref{sec:dqn} and Section~\ref{sec:dqn_dist}, respectively). In the end, the theoretical analysis of the proposed mechanism is provided (Section~\ref{sec:thm}).

\subsection{Vector-State Dynamics Model}
\label{sec:vector_model}

Our goal is not to perform multi-step rollout or long-horizon planning, but to obtain a lightweight, action-conditional signal that reflects how different actions are expected to unfold in the next step. In environments where the state is represented as a low-dimensional vector, this can be achieved by modeling the dynamics directly in the original state space.

We therefore learn a stochastic dynamics model $\mathcal{M}$ that predicts both the successor state distribution $P(s' \mid s, a)$ and the immediate reward. For state transitions, we model the dominant stochasticity using a Gaussian distribution,
\begin{equation}
    P(s' \mid s, a) \sim \mathcal{N}(\mu(s, a), \sigma^2(s, a)),
\end{equation}
where the mean $\mu(s, a)$ and variance $\sigma^2(s, a)$ are parameterized by neural networks $f_\mu$ and $f_\sigma$:
\begin{align}
    \mu(s, a) &= f_\mu(s, a), \\
    \sigma^2(s, a) &= \exp\!\left( f_\sigma(s, a) \right).
\end{align}
Given a state--action pair $(s,a)$, a successor state $s'_{\mathcal{M}}$ is sampled using the reparameterization trick,
\begin{equation}
    s'_{\mathcal{M}} = \mu(s, a) + \sigma(s, a)\cdot \epsilon, 
    \quad \epsilon \sim \mathcal{N}(0, I),
    \label{eq:successor_state}
\end{equation}
and the immediate reward is predicted by a separate head,
\begin{equation}
    r_{\mathcal{M}} = f_r(s, a).
\end{equation}
The dynamics model is trained using supervised losses on both the successor state and the reward,
\begin{equation}
    \mathcal{L}_{\mathcal{M}} 
    = \text{MSE}(s'_{\mathcal{M}}, s') 
    + \text{MSE}(r_{\mathcal{M}}, r),
    \label{eq:m_loss}
\end{equation}
where $(s,a,r,s')$ are sampled from the replay buffer. Once trained, $\mathcal{M}$ provides one-step rollout predictions $(s'_{\mathcal{M}}, r_{\mathcal{M}})$ that serve as auxiliary signals when forming the TD target, complementing the maximized Q-value.

\subsection{Image-State Dynamics Model}
\label{sec:image_model}

In image-based environments, the system state is not directly observable and must be inferred from high-dimensional observations. To obtain one-step successor predictions, we operate in a learned latent space. While recurrent state-space models (RSSMs)~\citep{hafner2019learning} are commonly used for long-horizon dynamics modeling, our objective here is more limited. We require a compact latent representation that supports reliable one-step prediction for TD target aggregation, rather than multi-step imagination or planning.

Accordingly, we adopt a simplified RSSM-style latent dynamics model focusing exclusively on immediate transition consequences. The model takes the current latent state $x_s$ and action $a$ as input, and produces predictions of the successor latent state and reward. By restricting the temporal scope to a single step, we avoid introducing recurrent structure that would add parameterization without providing corresponding benefit for our setting.
Specifically, given the one-hot action embedding $e_a$, the latent dynamics are defined as
\begin{equation}
\begin{aligned}
\textit{Feature model:} \quad 
& h = f_\phi(x_s, e_a), \\
\textit{Encoder:} \quad 
& z \sim q_\phi(z \mid h, x_{s'}), \\
\textit{Dynamics predictor:} \quad 
& \hat{z} \sim p_\phi(\hat{z} \mid h), \\
\textit{Reward predictor:} \quad 
& \hat{r} \sim p_\phi(\hat{r} \mid h, z), \\
\textit{Decoder:} \quad 
& \hat{x}_{s'} \sim p_\phi(\hat{x}_{s'} \mid h, z).
\end{aligned}
\label{eq:latent_model}
\end{equation}
The encoder $q_\phi(z \mid h, x_{s'})$ and the dynamics predictor $p_\phi(\hat{z} \mid h)$ together capture the stochastic transition structure in the latent space, while the decoder maps latent predictions back to the observation space. A separate reward head predicts the immediate reward associated with the transition. Importantly, this model is trained and used strictly for one-step prediction, and is not employed for trajectory rollout or policy optimization.

The dynamic model parameters $\phi$ are learned end-to-end using a weighted combination of prediction, dynamics, and representation losses, following the general structure of Dreamer-style training~\citep{hafner2025mastering}:
\begin{equation}
\mathcal{L}(\phi)=
\beta_{\text{pred}}\,\mathcal{L}_{\text{pred}}+
\beta_{\text{dyn}}\,\mathcal{L}_{\text{dyn}}+
\beta_{\text{rep}}\,\mathcal{L}_{\text{rep}},
\label{eq:wm_loss}
\end{equation}
where we set $\beta_{\text{pred}}=1$, $\beta_{\text{dyn}}=1$, and $\beta_{\text{rep}}=0.1$. The individual loss terms are given by
\begin{equation}
\begin{aligned}
\mathcal{L}_{\text{pred}}(\phi)
&=
\mathcal{L}_{\text{recon}}(\phi)
+ \mathcal{L}_{\text{rew}}(\phi)\\
&=
- \ln p_\phi(x_{s'} \mid h, z)
- \ln p_\phi(r \mid h, z), \\
\mathcal{L}_{\text{dyn}}(\phi)
&=
\max \Big(
1,\;
\mathrm{KL}
\big[
\mathrm{sg}\!\left(q_\phi(z \mid h, x_{s'})\right)
\,\big\|\,
p_\phi(z \mid h)
\big]
\Big), \\
\mathcal{L}_{\text{rep}}(\phi)
&=
\max \Big(
1,\;
\mathrm{KL}
\big[
q_\phi(z \mid h, x_{s'})
\,\big\|\,
\mathrm{sg}\!\left(p_\phi(z \mid h)\right)
\big]
\Big).
\end{aligned}
\end{equation}
To reduce the computational overhead of dynamics model training, we adapt the update frequency of the dynamics model based on the reconstruction component of \(\mathcal{L}_{\text{pred}}\), controlled by a threshold \(\tau_{\mathcal{M}}\). Specifically, if
\begin{equation}
    \mathcal{L}_{\text{recon}}(\phi) = -\ln p_\phi(x_{s'} \mid h, z) < \tau_{\mathcal{M}},
\end{equation}
the update frequency is reduced to a fraction $t_f \in (0, 1)$ of the default schedule; otherwise, the dynamics model follows the default update schedule. This adaptive schedule preserves frequent updates when the model has not yet learned reliable observation reconstruction, while reducing redundant updates after the reconstruction quality becomes sufficiently accurate.

Once the dynamics model is trained, the latent dynamics model provides one-step successor state and reward predictions that are used solely to inform TD target aggregation. The model does not alter the underlying Q-learning update, but supplies auxiliary predictive structure that helps regularize action comparison under uncertain value estimates.

\subsection{Structural Aggregation Target}
\label{sec:dqn}

In standard Q-learning, the TD target is formed by aggregating future action values through a hard maximization over next-state actions. Rather than modifying the off-policy learning framework itself, we focus on how this aggregation is carried out in the bootstrapped update. Our goal is to introduce additional structure into the comparison among candidate next actions, while leaving the underlying value function approximation unchanged.
To this end, we use the learned dynamics model $\mathcal{M}$ to obtain one-step, action-conditional predictions from the observed next state $s'$. For each candidate action $a' \in \mathcal{A}$, the model produces a predicted successor state $s''_{\mathcal{M}}(a')$ together with an immediate reward estimate. These predictions are not used for planning or policy improvement, but serve as auxiliary evidence when comparing candidate actions in the TD target.

We begin by recalling the Bellman optimality equation for the action-value function,
\begin{equation}
Q^*(s,a)=\mathbb{E}\!\left[r + \gamma \max_{a'} Q^*(s',a')\mid s,a\right].
\label{eq:bellman_opt}
\end{equation}
In practice, Q-learning approximates this relation using the greedy TD target $\max_{a'} Q'_\phi(s', a')$. As discussed earlier, the reliability of this update depends entirely on the quality of the maximized estimate at the observed next state $s'$.

To regularize this aggregation step, we additionally evaluate the immediate consequences of each candidate next action using the dynamics model $\mathcal{M}_\theta$. For every $a' \in \mathcal{A}$, the model generates a one-step rollout,
\begin{equation}
(r'_{\mathcal M}, s''_{\mathcal M})
\sim
\mathcal{M}_\theta(\cdot \mid s', a'),
\label{eq:rollout}
\end{equation}
from which we compute a model-based Bellman estimate,
\begin{equation}
\widetilde{Q'}_\phi(s', a')
=
r'_{\mathcal M}
+
\gamma
\max_{a''} Q'_\phi(s''_{\mathcal M}, a'').
\label{eq:model_bellman_correct}
\end{equation}
This quantity provides a rollout-informed evaluation of action $a'$, grounded in its predicted transition outcome rather than solely in the value estimate at $s'$.

We use these rollout-based estimates only to guide action comparison. Specifically, we select
\begin{equation}
\hat{a}'
=
\arg\max_{a' \in \mathcal{A}}
\widetilde{Q'}_\phi(s', a'),
\label{eq:a_q}
\end{equation}
and form a mixed TD target,
\begin{equation}
\hat{y}
=
\alpha
\max_{a'} Q'_\phi(s', a')
+
(1-\alpha)
Q'_\phi(s', \hat{a}'),
\label{eq:mixed_target_correct}
\end{equation}
where $\alpha \in [0,1]$ controls the interpolation between the standard greedy evaluation and the model-guided action selection. Importantly, the rollout does not replace the value function in the target; it only influences which action is used when forming the update. In this way, SADQ separates action ranking from value evaluation, reducing sensitivity to unreliable maxima while preserving the standard bootstrap structure.

\subsection{Extension to Distributional DQN}
\label{sec:dqn_dist}

We next extend the proposed aggregation mechanism to the distributional setting, where the target network predicts a return distribution $Z'_\phi(s,a)$ rather than a scalar value. In distributional DQN, greedy evaluation is typically performed with respect to the expected return,
\begin{equation}
\max_{a'} \mathbb{E}\!\left[ Z'_\phi(s', a') \right],
\end{equation}
and the TD target is formed by selecting the action with the largest expectation.

As in the scalar case, our objective is not to alter the underlying distributional learning framework, but to regularize how candidate actions are compared when forming the target. To this end, we use the learned dynamics model $\mathcal{M}_\theta$ to obtain one-step, action-conditional successor predictions from the observed next state $s'$. For each $a' \in \mathcal{A}$, the model produces
\begin{equation}
(r'_{\mathcal M}, s''_{\mathcal M})
\sim
\mathcal{M}_\theta(\cdot \mid s', a').
\end{equation}
Based on these predictions, we construct a rollout-informed distributional Bellman estimate,
\begin{equation}
\widetilde{Z'}_\phi(s', a')
=
r'_{\mathcal M}
+
\gamma \, Z'_\phi(s''_{\mathcal M}, a^{**}),
\label{eq:model_bellman_dist}
\end{equation}
where
\(
a^{**}
=
\arg\max_{a''} \mathbb{E}\!\left[ Z'_\phi(s''_{\mathcal M}, a'') \right].
\)
This estimate reflects the predicted immediate transition outcome of action $a'$, while retaining the distributional representation of future returns.

We use the expectation of these rollout-informed distributions to guide action comparison,
\begin{equation}
\hat a'
=
\arg\max_{a' \in \mathcal{A}}
\mathbb{E}\!\left[ \widetilde{Z'}_\phi(s', a') \right],
\label{eq:a_q_dist}
\end{equation}
and construct a mixed distributional target,
\begin{equation}
\hat Z
=
\alpha \, Z'_\phi(s', a^*)
+
(1-\alpha) \, Z'_\phi(s', \hat a'),
\label{eq:distributional_mixed_target}
\end{equation}
where
\(
a^*
=
\arg\max_{a'} \mathbb{E}\!\left[ Z'_\phi(s', a') \right].
\)

This formulation preserves the same structural aggregation principle as in the scalar case. The dynamics model is used only to influence which action is selected for the target, while the value distribution itself is always provided by the distributional critic. In this way, the proposed mechanism extends naturally from scalar Q-values to return distributions, regularizing greedy action comparison without modifying the distributional update rule.

\subsection{Theoretical Analysis}
\label{sec:thm}
We provide a theoretical analysis to clarify how the proposed aggregation affects bootstrap bias, and to verify that it does not alter the optimal solution asymptotically. All formal proofs are deferred to \textbf{Appendix~\ref{appx:theory_proof}}.

Let $\mathcal{M}_\theta(\cdot \mid s,a)$ denote a dynamics model trained on the support of the replay buffer $\mathcal{D}$. For notational convenience, define the standard and model-based TD targets
\begin{align}
y_{\mathcal M} &= r_{\mathcal M} + \gamma \max_{a'} Q'_\phi(s'_{\mathcal{M}}, a'), \\
y &= r + \gamma \max_{a'} Q'_\phi(s', a'),
\end{align}
where $(r,s') \sim \mathcal{T}(\cdot \mid s,a)$ follows the true environment dynamics and $(r_{\mathcal M}, s'_{\mathcal M}) \sim \mathcal{M}_\theta(\cdot \mid s,a)$ is generated by the learned dynamics model.

\paragraph{Lemma 1 (Local state extrapolation).}
Under the neural tangent kernel (NTK) regime~\citep{jacot2018neural}, for any in-sample state-action pair $(s,a) \in \mathcal{D}$ and in-neighborhood state-action pair $(s_{\mathcal{M}}, a)$ such that $\|s - s_{\mathcal{M}} \| \leq \epsilon$, the value difference of the deep Q function can be bounded as:
\begin{equation}\label{eq:lemma1}
\begin{aligned}
    \|Q_{\phi}&(s'_{\mathcal M},a') - Q_{\phi}(s',a')\| \leq \\ 
    & C\left(\sqrt{\min\left(\|s'\oplus a'\|,\|s'_{\mathcal M}\oplus a'\|\right)}\sqrt{\epsilon}+ 2\epsilon \right),
\end{aligned}
\end{equation}
where $\oplus$ denotes the vector concatenation operation, and C is a finite constant. The \textbf{Lemma~1} is a direct corollary of Theorem 1 in \cite{li2023when}, specialized to the case of local state extrapolation.

\paragraph{Assumption 1 (Controlled model-induced perturbation).}
We assume that for all $(s,a) \in \mathcal{D}$,
\begin{equation}
\mathbb{E}\!\left[
\| y_{\mathcal M} - y \|
\,\middle|\, (s,a)
\right]
\;\le\;
\epsilon_r + \gamma \epsilon_s,
\label{eq:assum}
\end{equation}
where $\epsilon_r$ bounds the reward prediction error of the learned model, and $\epsilon_s$ bounds the state-induced value perturbation characterized by \textbf{Lemma~1}, respectively.

\paragraph{Theorem 1 (Bootstrap bias reduction and ideal-limit target consistency).}
Let $\mathcal{T}_{\mathrm{std}}$ and $\mathcal{T}_{\mathrm{mix}}$ denote the standard Bellman operator and the mixed operator induced by Eq.~(\ref{eq:mixed_target_correct}), respectively. Then, the following properties hold:
\begin{enumerate}
    \item For any bounded action-value function $Q$, the mixed operator satisfies
    \[
    \mathcal{T}_{\mathrm{mix}} Q \;\le\; \mathcal{T}_{\mathrm{std}} Q
    \quad \text{pointwise}.
    \]
    \item In the ideal limiting case where the model-induced perturbation approach zero in \textbf{Assumption 1}, then
    \[
    \lim_{k \to \infty}
    \big\|
    \mathcal{T}_{\mathrm{mix}}^{(k)} Q^* - \mathcal{T}^* Q^*
    \big\|_\infty
    = 0.
    \]
\end{enumerate}
The first property shows that, for any given $Q$, the mixed operator produces a TD target no larger than the standard greedy backup. This indicates that SADQ attenuates the upward bias caused by unreliable maximized value estimates. The second property is an ideal-limit target consistency result. When the model-induced perturbation approaches zero, the mixed target approaches the optimal Bellman target. This does not imply a finite-error fixed-point guarantee; with nonzero residual model error, the induced operator may differ from the optimal Bellman operator.

\begin{algorithm}[t]
    \caption{SADQ}
    \label{alg:sadq}
    \textbf{Input}: Online Q-network \(Q(s, a; \theta)\), target Q-network \\
    \(Q(s, a; \theta')\), replay buffer \(\mathcal{D}\), dynamics model \(\mathcal{M}\). \\
    \textbf{Output}: Optimized Q-network \(Q(s, a; \theta^*)\). \\
    \begin{algorithmic}[1]
        \STATE \textbf{Training loop:}
        \WHILE{not done}
            \STATE Sample transitions \(\{(s, a, r, s')\}\) from \(\mathcal{D}\) for \(\mathcal{M}\).
            \STATE Train model \(\mathcal{M}\) by Eq.~(\ref{eq:m_loss}) or Eq.~(\ref{eq:wm_loss}).
            \STATE Resample transitions \(\{(s, a, r, s')\}\) from \(\mathcal{D}\) for Q.
            \STATE Compute successor rollout in Eq.~(\ref{eq:rollout}).
            \STATE Get candidate action \(\hat{a}'\) in Eq.~(\ref{eq:a_q}) or Eq.~(\ref{eq:a_q_dist}).
            \STATE Compute target Q-value \(y\) in Eq.~(\ref{eq:mixed_target_correct}) or Eq.~(\ref{eq:distributional_mixed_target}).
            \STATE Update Q-networks.
            \STATE Periodically update target network \(Q'\).
        \ENDWHILE
        \STATE \textbf{return} Optimized Q-network parameters \(\theta\).
    \end{algorithmic}
\end{algorithm}

\begin{figure*}[ht]
    \centering
    \subfloat[]{
        \includegraphics[width=0.23\textwidth]{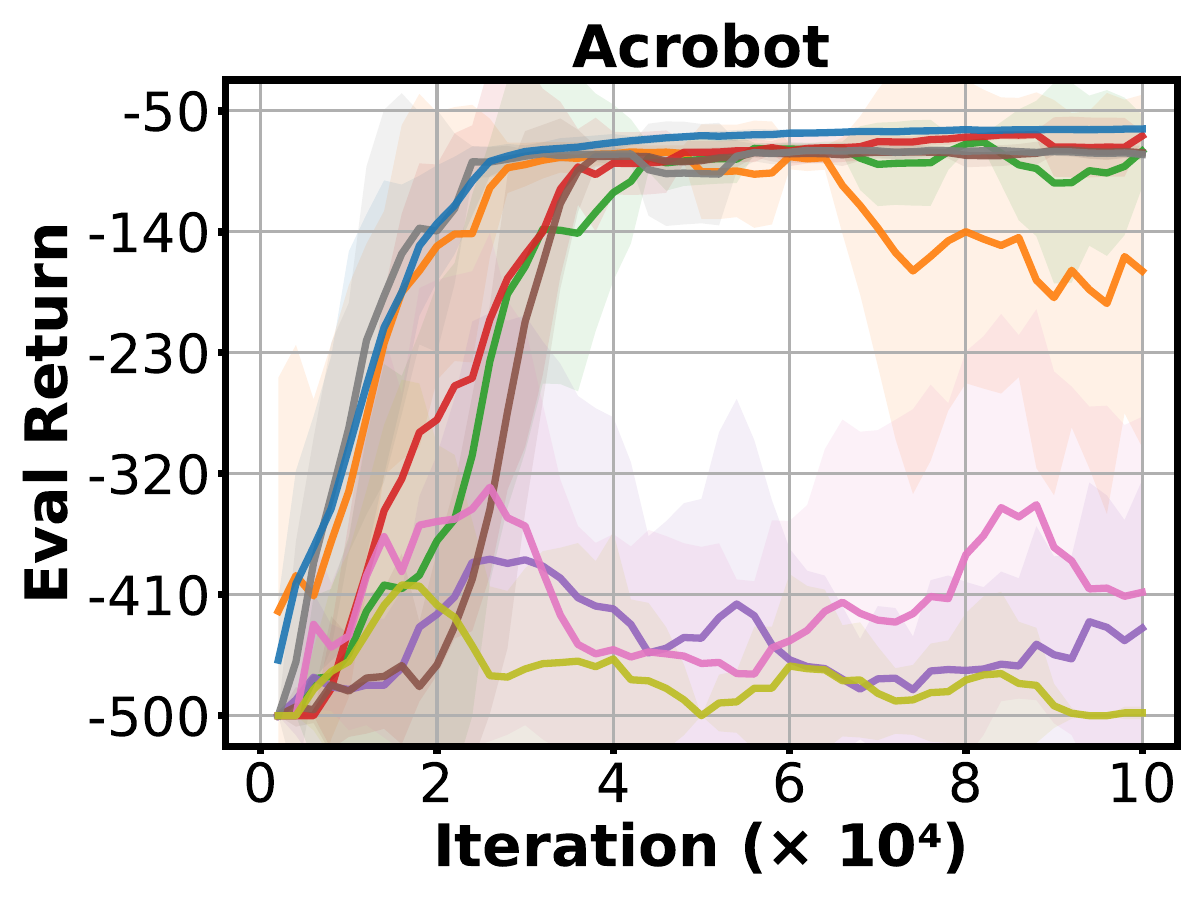}
        \label{fig:acrobot}
    }
    \hfill
    \subfloat[]{
        \includegraphics[width=0.23\textwidth]{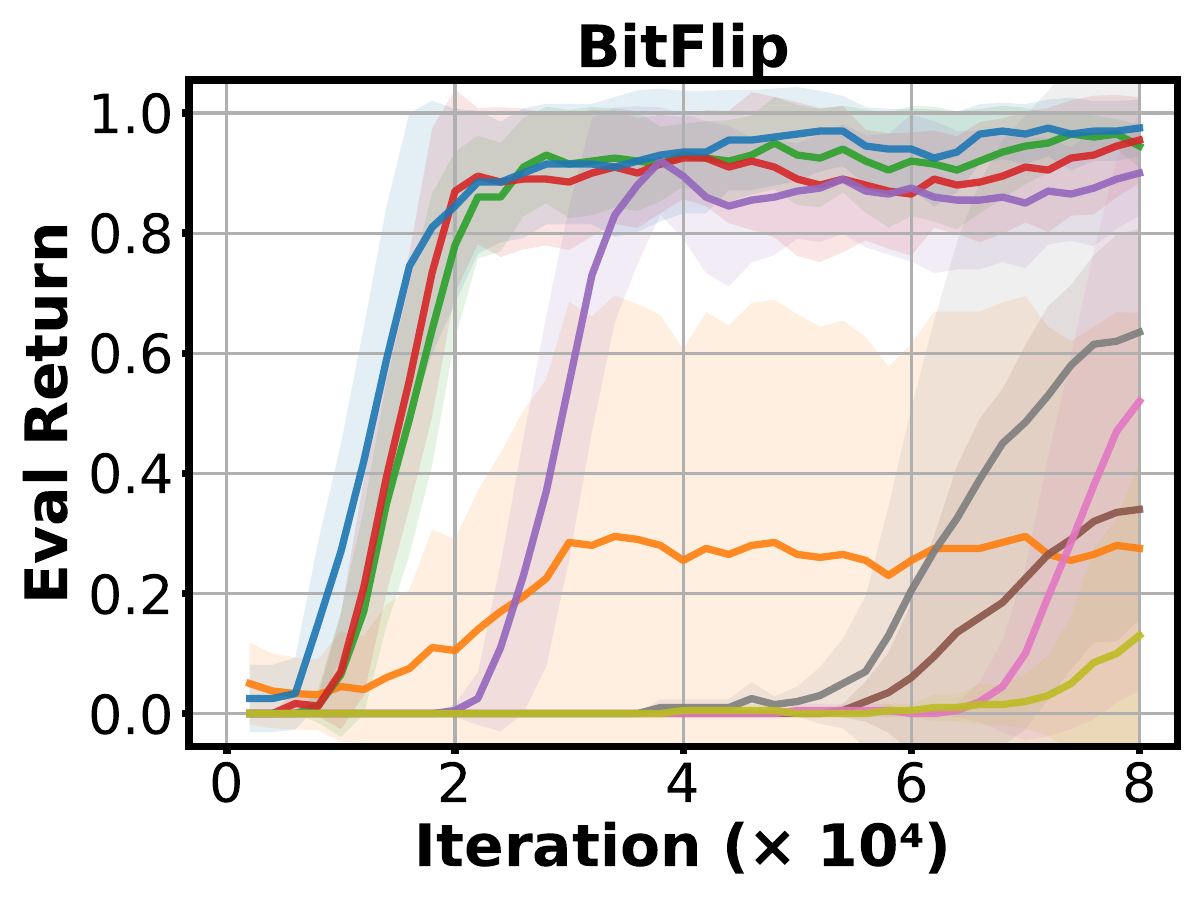}
        \label{fig:bitflip}
    }
    \hfill
    \subfloat[]{
        \includegraphics[width=0.23\textwidth]{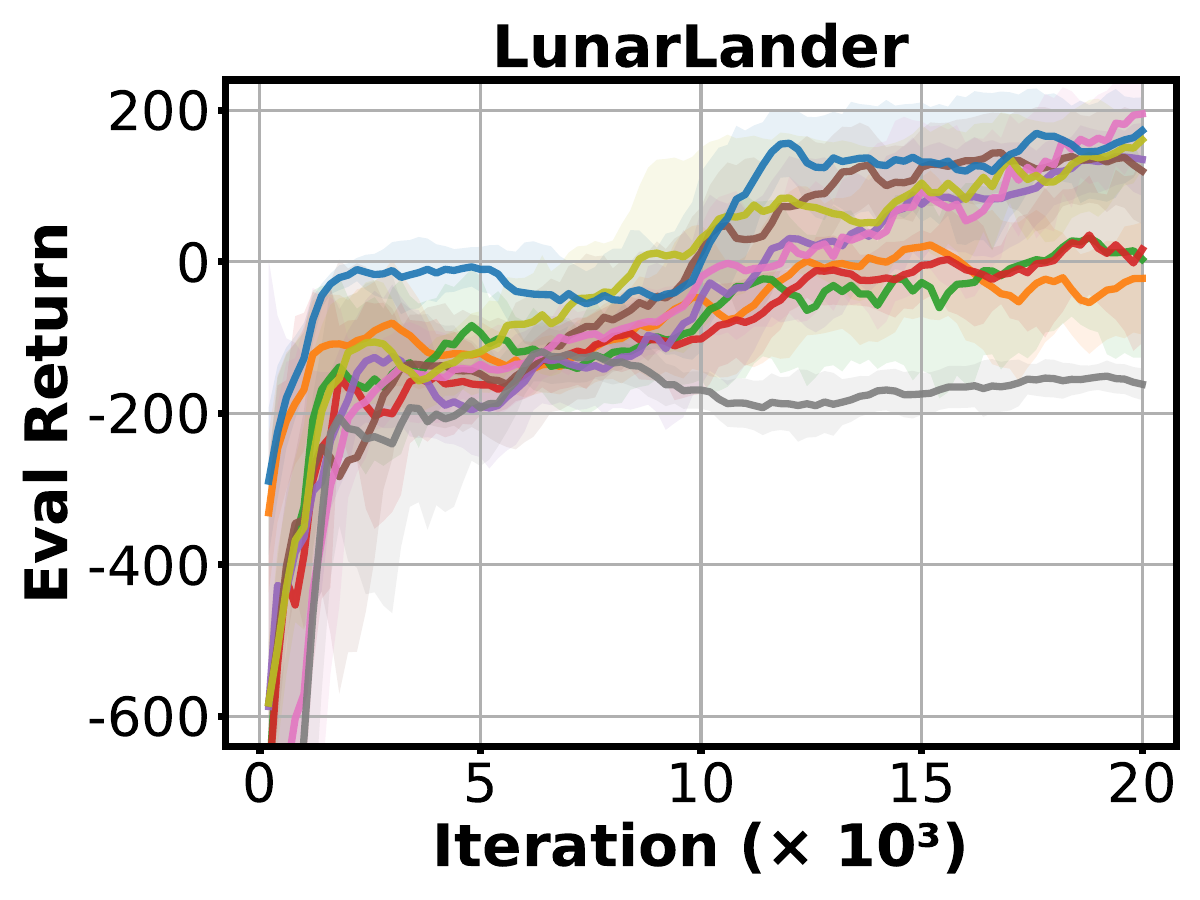}
        \label{fig:lunarlander}
    }
    \hfill
    \subfloat[]{
        \includegraphics[width=0.23\textwidth]{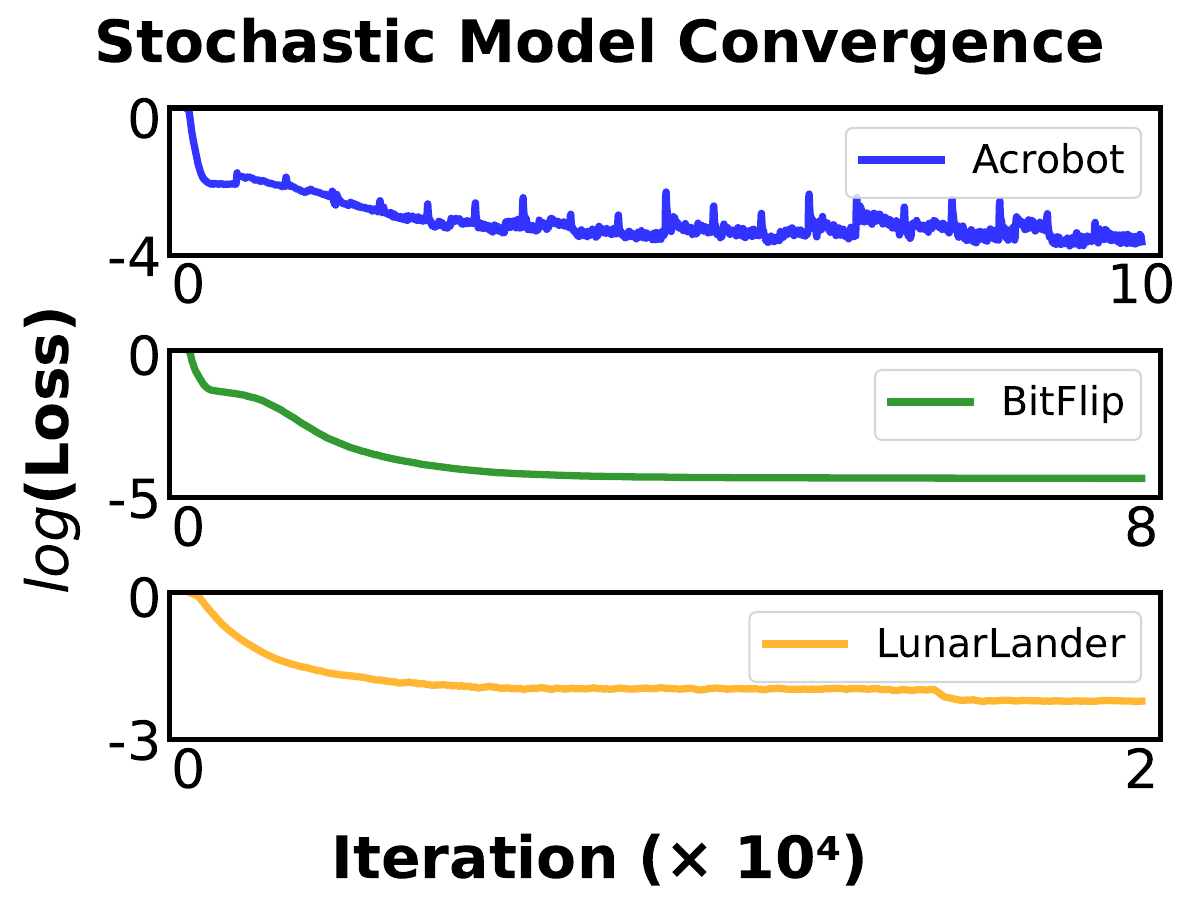}
        \label{fig:prob_loss}
    }
    \hfill
    \includegraphics[width=0.8\textwidth]{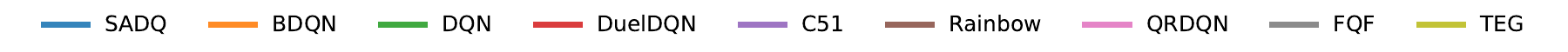}  
    \caption{Performance comparison of SADQ with other baselines across conventional RL tasks. The legend above describes the corresponding methods in (a) Acrobot, (b) BitFlip, (c) LunarLander, and (d) Convergency of the stochastic model.}
    \label{fig:main}
\end{figure*}

\begin{figure}[t]
    \centering
    \subfloat[]{
        \includegraphics[width=0.22\textwidth]{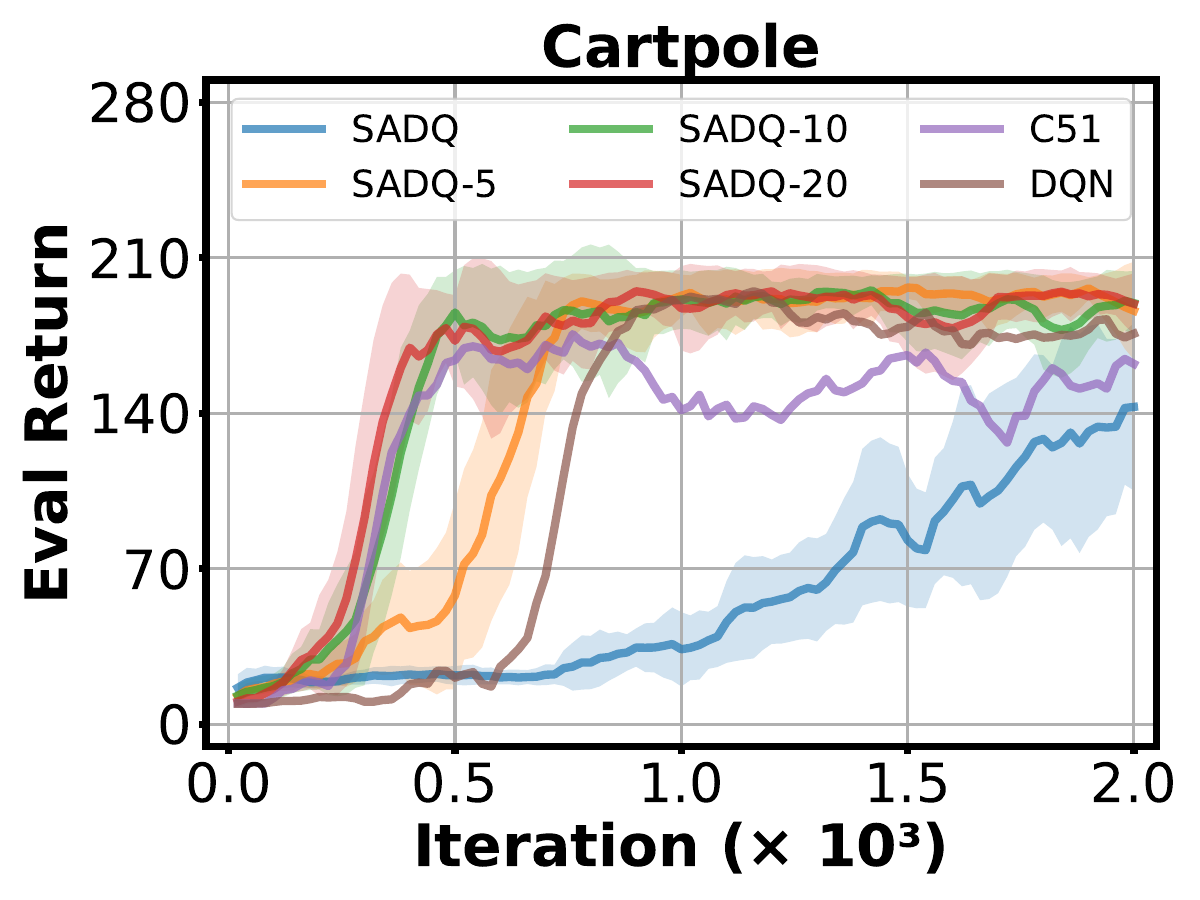}
        \label{fig:cartpole_train}
    }
    \hfill
    \subfloat[]{
        \includegraphics[width=0.22\textwidth]{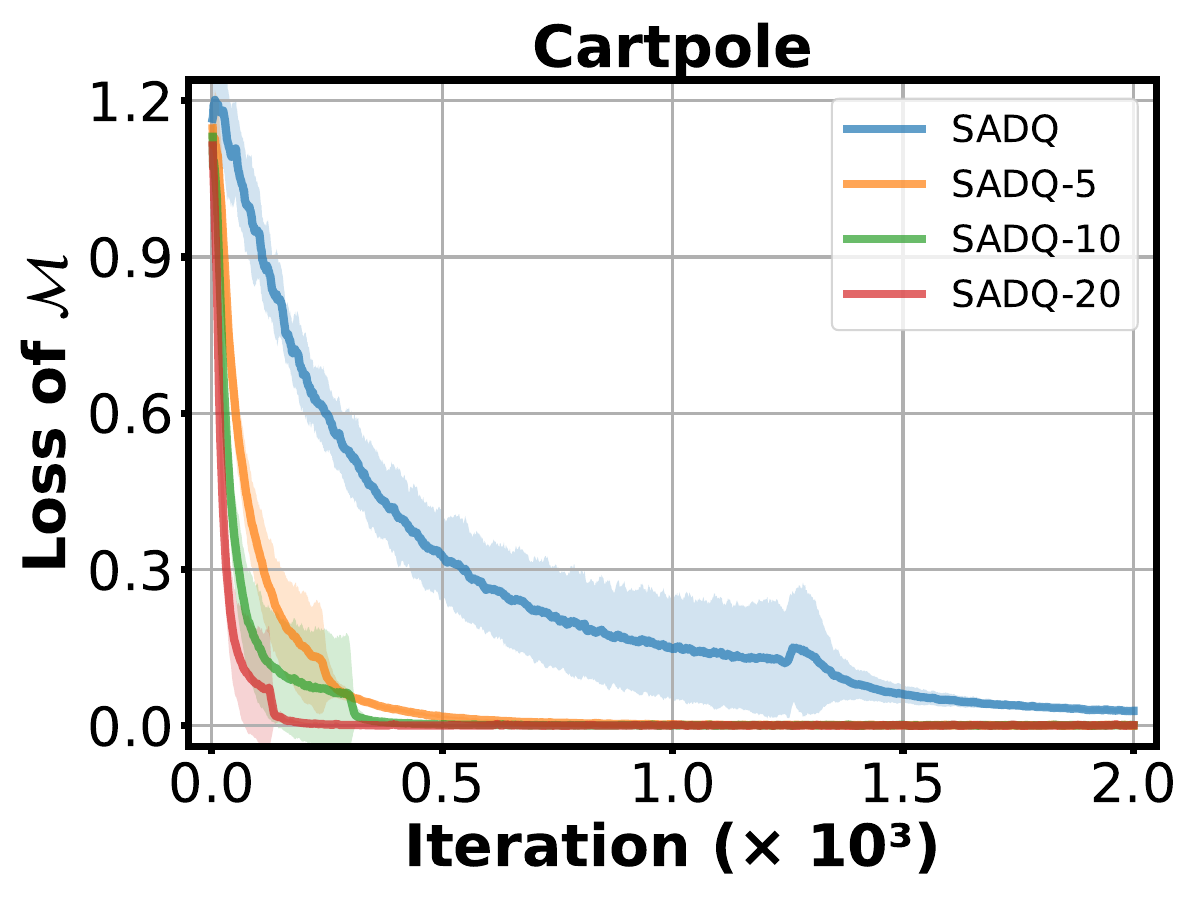}
        \label{fig:cartpole_prob}
    }
    \caption{Effects of stochastic model convergence to performance. (a) Effects of updating frequency to SADQ and compare with C51 and DQN; (b) The Loss of stochastic model among different updating configurations.}
    \label{fig:convergence}
\end{figure}

\subsection{SADQ Algorithm}
The Algorithm~\ref{alg:sadq} summarizes how SADQ modifies TD target construction in standard Q-learning. Instead of relying solely on the max operator over next-state Q-values, SADQ uses one-step successor rollout information from the auxiliary dynamics model to guide action aggregation. The Q-network is then updated with the resulting mixed TD target, while the remaining replay-based training pipeline follows the standard DQN-style procedure.

\section{Experiments}

We first evaluate SADQ on discrete-action vector-based benchmarks implemented in DI-Engine\footnote{\url{https://github.com/opendilab/DI-engine}}. In this setting, SADQ is built upon the Dueling DQN backbone~\citep{wang2016dueling}. The environments include Acrobot~\citep{sutton1995generalization}, BitFlip, Cartpole~\citep{barto1983neuronlike}, and LunarLander, along with two real-world-inspired scenarios, CityFlow~\citep{tang2019cityflow} and O-Cloud. Detailed environment descriptions are provided in \textbf{Appendix~\ref{appx:realworld}}.

We compare SADQ against representative value-based baselines from two major categories. \textbf{(1) Classical DQN variants:} DQN~\citep{mnih2015human}, Double DQN~\citep{van2016deep}, Dueling DQN~\citep{wang2016dueling}, BDQN~\citep{azizzadenesheli2018efficient}, and SUNRISE~\citep{lee2021sunrise}. \textbf{(2) Distributional DQN methods:} C51~\citep{bellemare2017distributional}, Rainbow~\citep{hessel2018rainbow}, QRDQN~\citep{dabney2018distributional}, FQF~\citep{yang2019fully}, and TEG~\citep{dabney2021temporallyextended}. This setup allows us to evaluate SADQ against both classical value estimators and distributional formulations.

To further assess generality in high-dimensional visual domains, we conduct experiments on the Atari100K benchmark~\citep{bellemare2013arcade}, implemented in Jax-baseline~\footnote{\url{https://github.com/tinker495/jax-baseline}}. In the image-based setting, SADQ is applied as a plug-in modification to multiple value-based agents. Specifically, we replace the standard TD target in QRDQN~\citep{dabney2018distributional}, IQN~\citep{dabney2018implicit}, and CoAct~\cite{korkmazcounteractive} with the proposed mixed target, while preserving their original architectures, distributional parameterizations, and optimization procedures. This design isolates the effect of SADQ at the target-construction level and enables evaluation across different quantile-based estimators.

All experiments are conducted with five independent random seeds. We report mean performance across seeds, and full hyperparameter configurations are provided in \textbf{Appendix~\ref{appx:experiments_config}} for reproducibility.

\subsection{Conventional RL Tasks}
We evaluate SADQ across multiple control environments. 
On Acrobot, BitFlip, and LunarLander (Figs.~\ref{fig:acrobot}--\ref{fig:lunarlander}), SADQ attains the highest or competitive final returns while exhibiting more stable training dynamics than representative DQN variants, which often show rapid early gains followed by performance degradation. To isolate the contribution of SADQ, we compare multiple baselines with and without SADQ under identical experimental settings, where the only difference is the max-operator-related component introduced by SADQ. As shown in \textbf{Appendix~\ref{appx:baseline_comparison}}, SADQ consistently improves all baselines.

To examine the role of the vector-based stochastic model $\mathcal{M}$, we vary its update frequency $k$ in Cartpole. As shown in Fig.~\ref{fig:cartpole_train}, larger $k$ consistently accelerates convergence and improves final performance, with $k=20$ achieving the fastest learning. The corresponding loss curves (Fig.~\ref{fig:cartpole_prob}) confirm faster and more stable model convergence as $k$ increases. Together, these results indicate that more accurate successor prediction leads to improved stability and effectiveness of the SADQ update. See \textbf{Appendix~\ref{appx:cartpole}} for further discussion of Cartpole.

\subsection{Real World Scenarios}
\begin{figure}[t]
    \centering
    \subfloat[]{
        \includegraphics[width=0.23\textwidth]{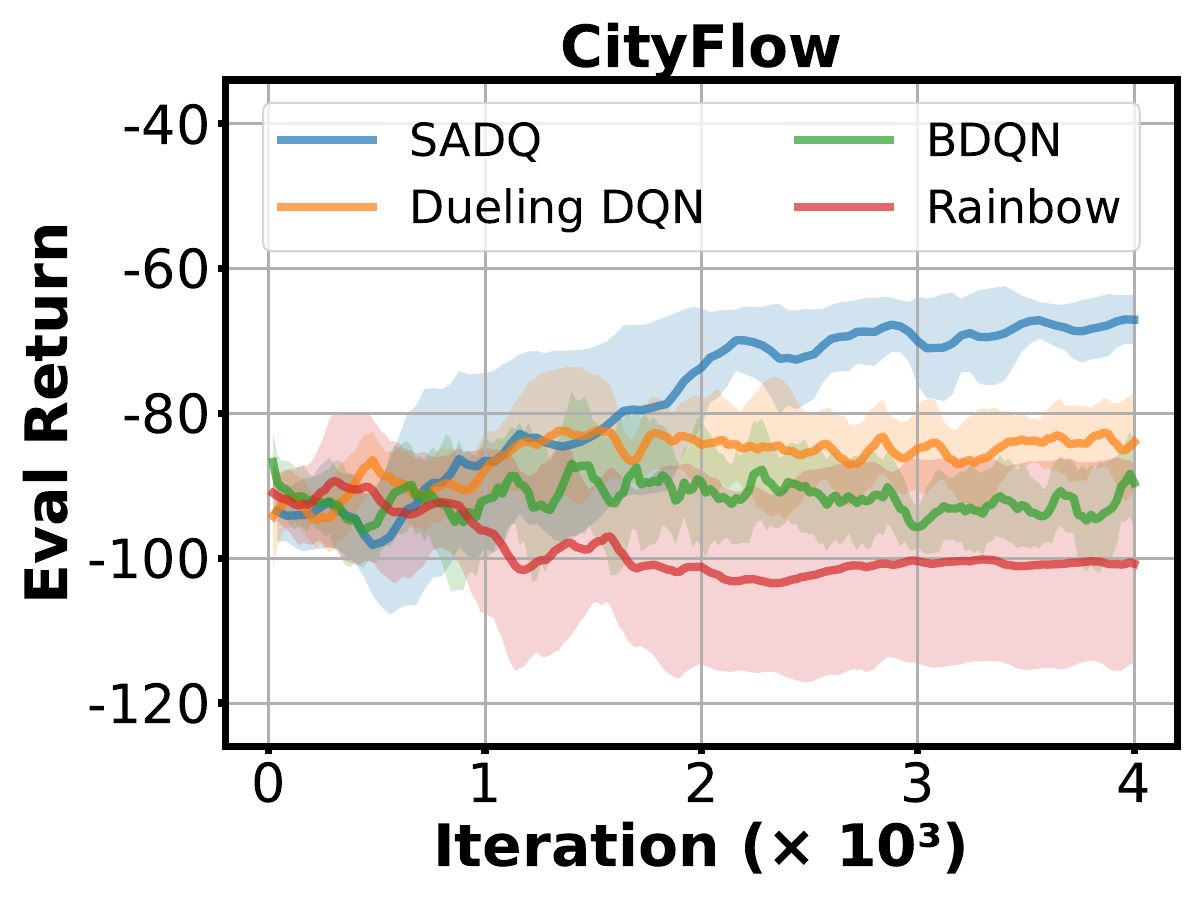}
        \label{fig:cityflow}
    }
    \hfill
    \subfloat[]{
        \includegraphics[width=0.23\textwidth]{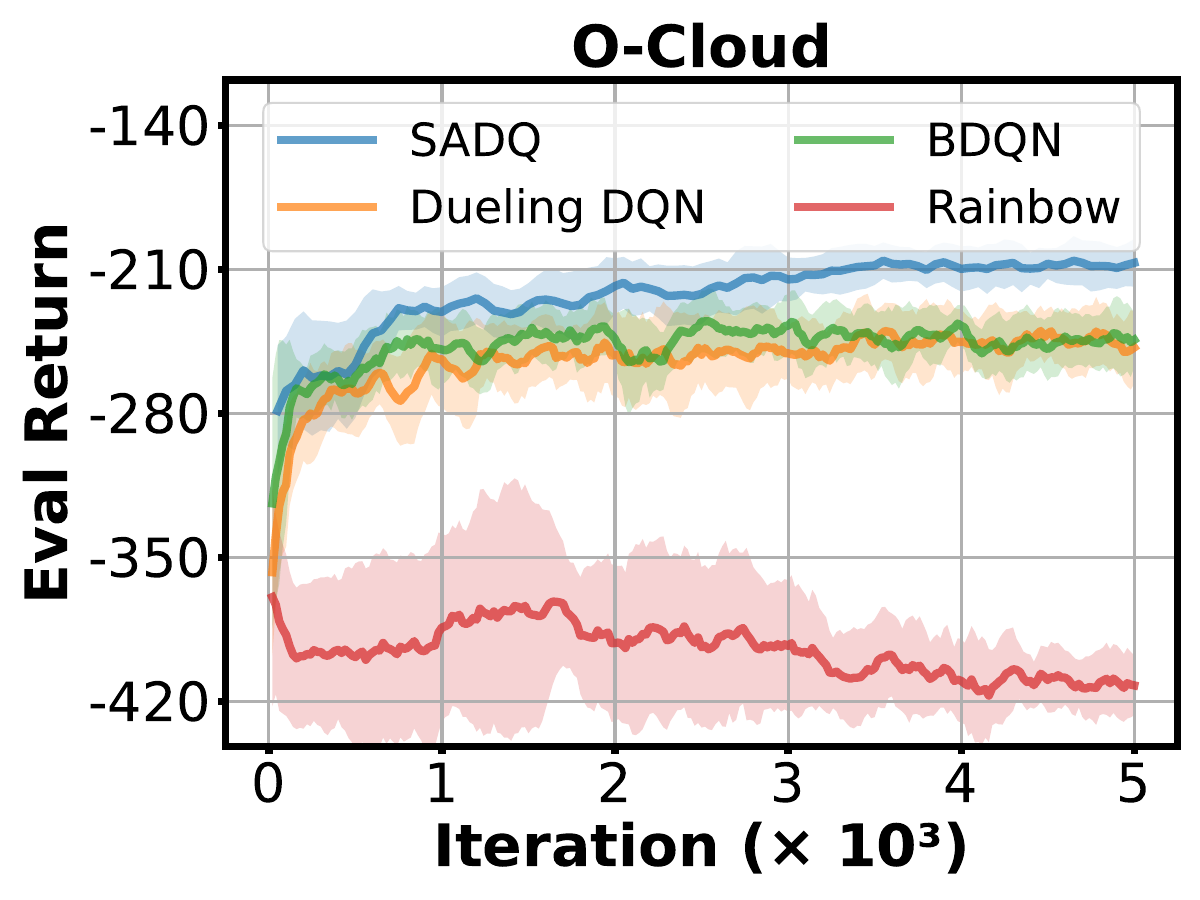}
        \label{fig:ocloud}
    }
    \caption{Performance comparison of SADQ with baselines across CityFlow and O-Cloud scenarios. (a) Performance in CityFlow scenario; (b) Performance in O-Cloud scenario.}
    \label{fig:real_world}
\end{figure}

We further evaluate SADQ in two real-world scenarios: CityFlow for traffic signal control and O-Cloud for dynamic resource allocation. These experiments are designed to assess the practical applicability of SADQ in complex, structured decision-making environments and to compare its performance against representative baseline methods under realistic operational conditions. In both scenarios (Figs.~\ref{fig:cityflow}-~\ref{fig:ocloud}), SADQ demonstrates superior performance compared to other baselines, including Dueling DQN, BDQN, and Rainbow. During the whole training process, SADQ achieves the highest evaluation return, consistently outperforming the baselines throughout training. 

\begin{figure}[ht]
    \centering
    \subfloat[]{
        \includegraphics[width=0.23\textwidth]{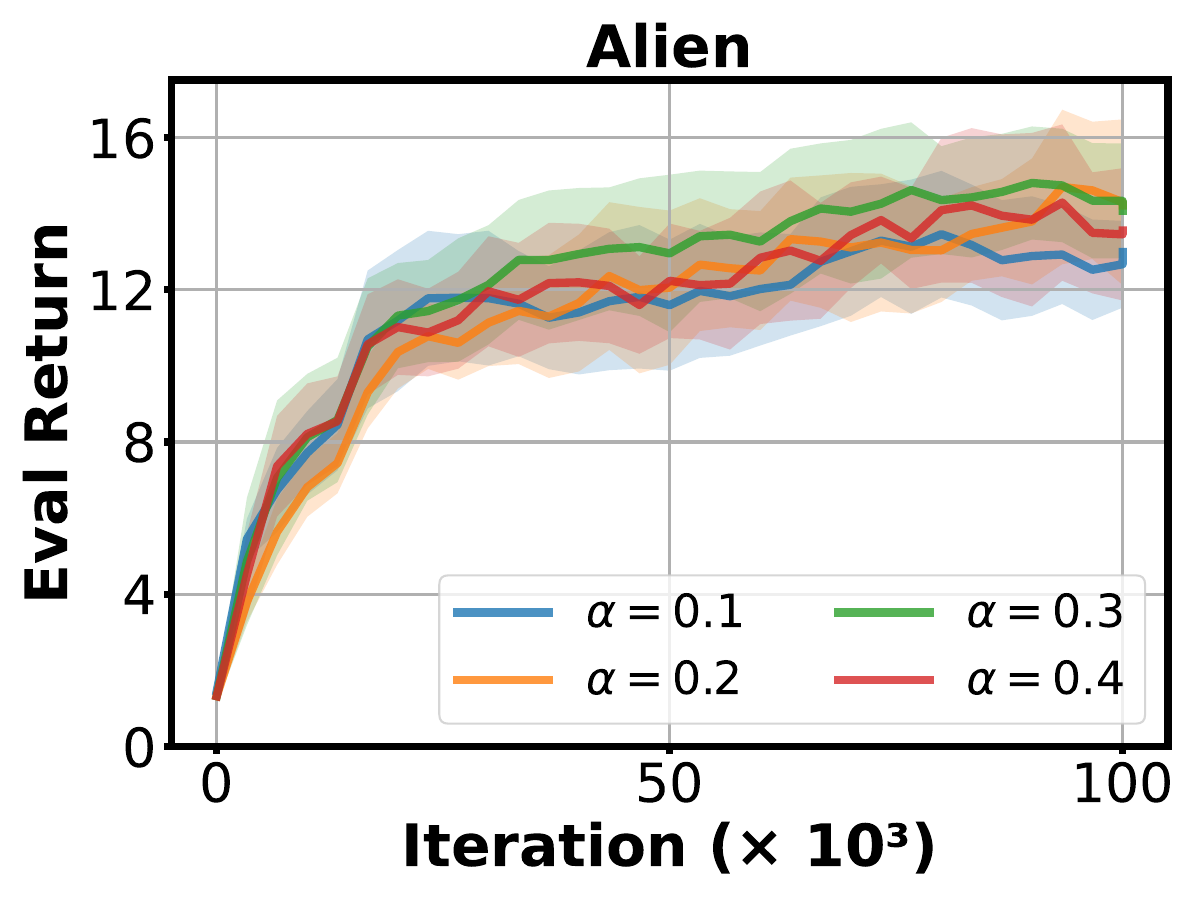}
        \label{fig:alpha_alien}
    }
    \hfill
    \subfloat[]{
        \includegraphics[width=0.23\textwidth]{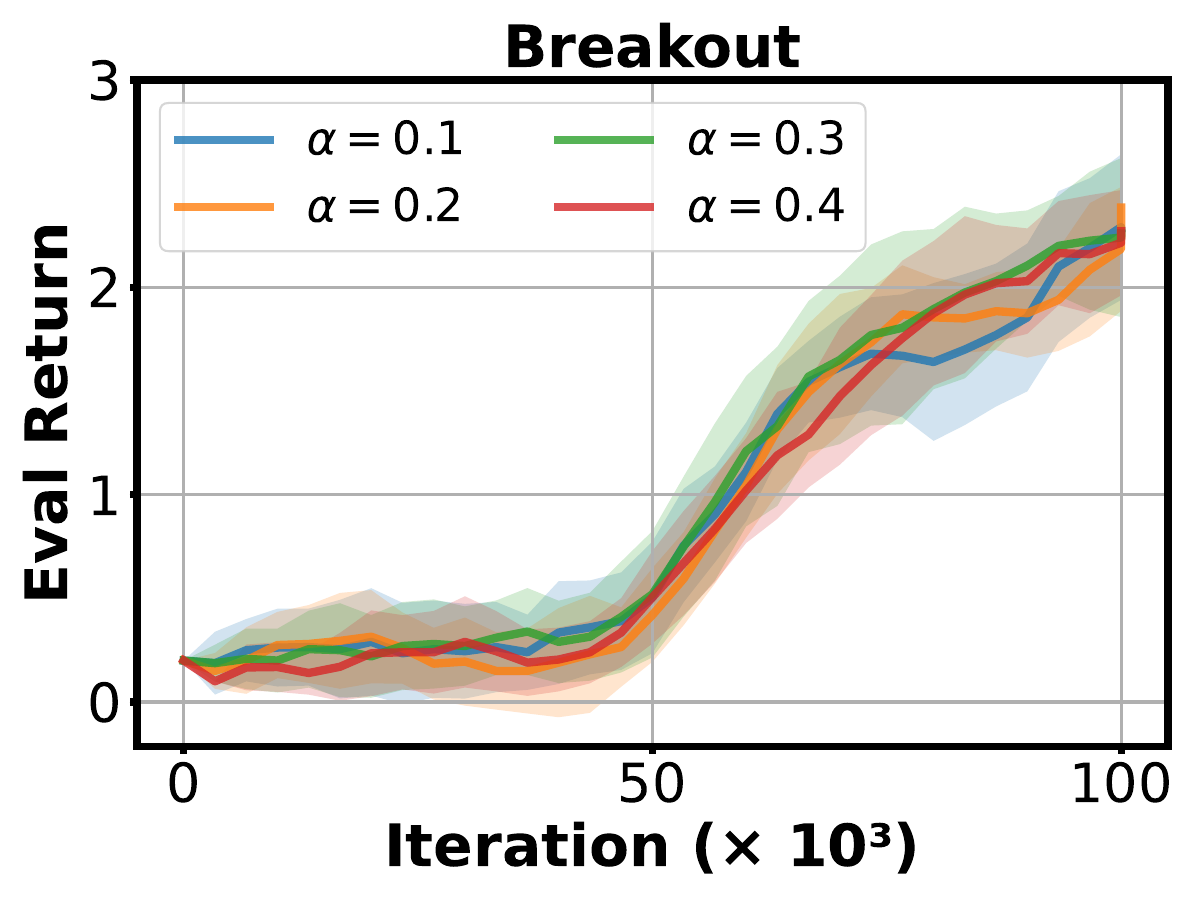}
        \label{fig:alpha_breakout}
    }
    \hfill
    \subfloat[]{
        \includegraphics[width=0.23\textwidth]{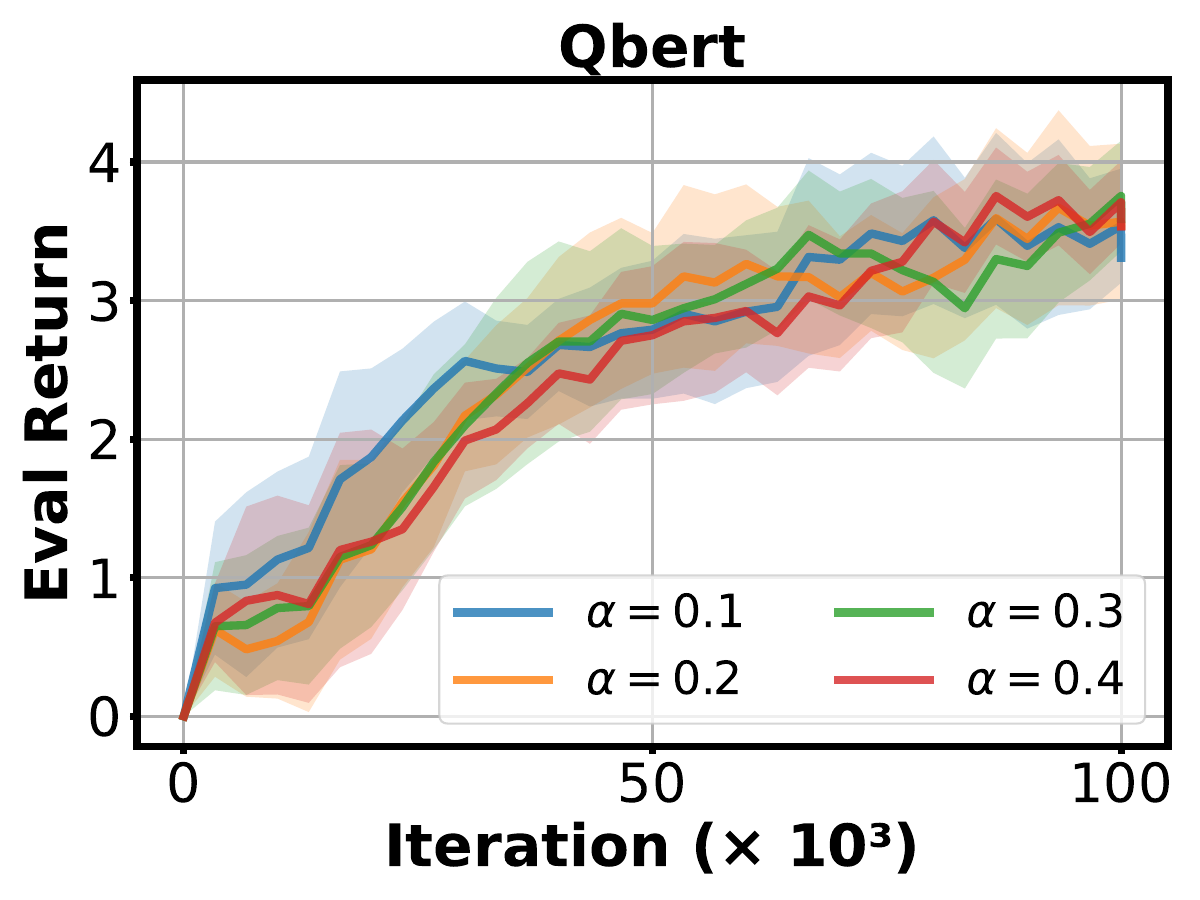}
        \label{fig:alpha_qbert}
    }
    \hfill
    \subfloat[]{
        \includegraphics[width=0.23\textwidth]{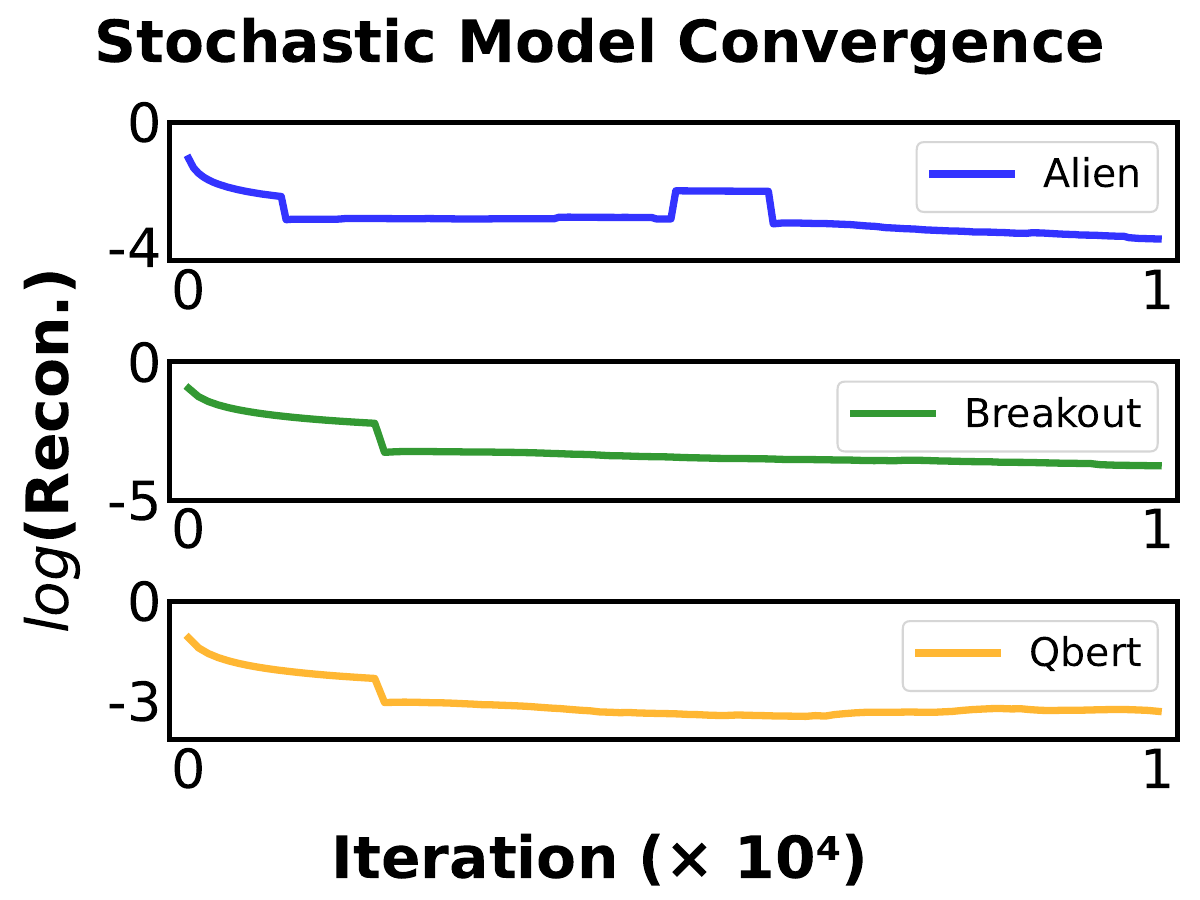}
        \label{fig:model_convergence}
    }
    \caption{Case study on Atari environments. (a) $\alpha$-ablation on Alien; (b) $\alpha$-ablation on Breakout; (c) $\alpha$-ablation on Qbert; (d) Convergence of the stochastic dynamics model measured by log reconstruction loss across three games.}
    \label{fig:atari_case_study}
\end{figure}

\begin{table*}[ht]
\centering
\caption{Performance on Atari100K. Normalized scores are averaged over 20 evaluations per seed.}
\label{tab:atari100k_full}
\begin{tabular}{l|cc|cc|cc}
\toprule
 & \multicolumn{2}{c|}{QRDQN} 
 & \multicolumn{2}{c|}{IQN} 
 & \multicolumn{2}{c}{CoAct} \\
\cline{2-7}
Game & Base & +SADQ & Base & +SADQ & Base & +SADQ \\
\midrule
Alien         & 12.49 & \textbf{14.22} & 10.08 & \textbf{10.98} & 10.74 & \textbf{11.54} \\
Amidar        & 2.48 & \textbf{3.15} & 2.61 & \textbf{3.04} & 2.06 & \textbf{2.85} \\
Assault       & 4.82 & \textbf{4.86} & \textbf{4.80} & 4.63 & 3.79 & \textbf{4.03} \\
Asterix       & 2.27 & \textbf{2.45} & 1.62 & \textbf{1.74} & 1.75 & \textbf{2.07} \\
BankHeist     & \textbf{0.30} & 0.23 & 0.08 & 0.08 & 0.15 & \textbf{0.21} \\
BattleZone    & 0.41 & \textbf{0.61} & 0.42 &\textbf{ 0.46} & 0.70 & \textbf{0.99} \\
Boxing        & 2.62 & \textbf{3.92} & -31.27 & \textbf{-26.91} & 2.84 & \textbf{3.43} \\
Breakout      & 2.00 & \textbf{2.38} & 2.30 & \textbf{2.31} & 1.38 & \textbf{1.43} \\
ChopperCommand& 1.83 & \textbf{2.02} & \textbf{1.63} & 1.37 & 1.80 & \textbf{1.88} \\
CrazyClimber  & 15.23 & \textbf{23.11} & 16.12 & \textbf{29.93} & 24.17 & \textbf{28.81} \\
DemonAttack   & 3.82 & \textbf{4.09} & \textbf{4.12} & 4.01 & \textbf{3.46} & 3.23 \\
Gopher        & 3.91 & \textbf{4.99} & 2.83 & \textbf{3.41} & 7.18 & \textbf{7.56} \\
Hero          & 1.63 & \textbf{3.30} & 1.06 & \textbf{2.08} & 3.09 & \textbf{6.40} \\
Jamesbond     & 0.16 & \textbf{0.24} & 0.18 & \textbf{0.22} & 0.17 & \textbf{0.18} \\
Kangaroo      & \textbf{0.26} & 0.22 & \textbf{0.04} & 0.02 & \textbf{0.35} & 0.24 \\
Krull         & 76.34 & \textbf{77.04} & 67.17 & \textbf{69.77} & 74.60 & \textbf{77.81} \\
KungFuMaster  & 7.70 & \textbf{8.93} & 1.60 & \textbf{4.65} & 1.74 & \textbf{2.05} \\
MsPacman      & 16.23 & \textbf{17.30} & 15.46 & \textbf{17.09} & 12.93 & \textbf{13.21} \\
Pong          & -14.96 & \textbf{-14.27} & -20.70 & -20.70 & -16.88 & \textbf{-16.66} \\
PrivateEye    & \textbf{-1.35} & -1.85 & -21.00 & \textbf{-8.46} & -4.91 & \textbf{-2.95} \\
Qbert         & 3.42 & \textbf{3.54} & 2.25 & \textbf{3.04} & 2.11 & \textbf{2.30} \\
RoadRunner    & \textbf{1.66} & 1.65 & 2.08 & \textbf{2.69} & 0.91 & \textbf{0.93} \\
Seaquest      & 2.80 & \textbf{3.22} & 1.84 & \textbf{2.59} & 2.48 & \textbf{2.63} \\
UpNDown       & 5.28 & \textbf{5.55} & 4.35 & \textbf{4.57} & 4.29 & \textbf{4.56} \\
\midrule
Mean          & 6.31 & \textbf{7.05} & 2.90 & \textbf{4.69} & 5.89 & \textbf{6.63} \\
\bottomrule
\end{tabular}
\end{table*}

\subsection{Atari100K Games}
Tab.~\ref{tab:atari100k_full} reports results on Atari100K games for QRDQN, IQN, and CoAct with and without SADQ. Across all three estimators, SADQ improves both mean and median performance, with gains on 20/24 games for QRDQN, 18/24 for IQN, and 22/24 for CoAct. These results demonstrate that successor-guided target formation consistently enhances robustness in quantile-based value learning. We further analyze the effect of the mixing coefficient $\alpha$ on three representative games (Alien, Breakout, and Qbert). As shown in Figs~\ref{fig:alpha_alien}--\ref{fig:alpha_qbert}, SADQ yields smoother learning dynamics and reduced variance for all tested $\alpha$, indicating robustness to the choice of mixing weight. Fig~\ref{fig:model_convergence} shows that the simplified RSSM-style model converges rapidly ($\sim$2k iterations), suggesting that the rollout signal becomes reliable early and effectively regularizes TD target construction.

In our implementation, $\mathcal{M}$ is instantiated as the simplified one-step RSSM-style model described in Sec.~\ref{sec:image_model}. The full RSSM contains approximately 62M parameters, whereas our one-step dynamics model contains approximately 34M parameters. To further examine the computational overhead of training the dynamics model $\mathcal{M}$, we report the wall-clock training time of SADQ under different dynamics-model update frequencies. Tab.~\ref{tab:runtime_overhead} compares the runtime with $t_f=40\%$ and $t_f=20\%$ on Atari-100K and Atari-10M. The results show that the adaptive scheduling mechanism maintains sufficient state-prediction accuracy while substantially reducing the computational overhead of dynamics-model training.

\begin{table}[H]
\centering
\caption{Runtime comparison of QR-DQN and SADQ with different reduced update fractions $t_f$.}
\label{tab:runtime_overhead}
\begin{tabular}{lccc}
\toprule
& \textbf{QR-DQN} & \textbf{+SADQ} & \textbf{+SADQ} \\
& & \textbf{($t_f=40\%$)} & \textbf{($t_f=20\%$)} \\
\midrule
Atari-100K & 4 mins   & 6 mins   & 5 mins   \\
Atari-10M  & 434 mins & 598 mins & 522 mins \\
\bottomrule
\end{tabular}
\end{table}

\section{Discussion}

SADQ relies on an auxiliary dynamics model $\mathcal{M}$ for one-step successor prediction, which introduces a trade-off between prediction reliability and computational overhead. If $\mathcal{M}$ has not sufficiently converged, its predictions may perturb action comparison, especially in simple tasks such as Cartpole where standard DQN-style updates already provide a strong learning signal. This trade-off comes from introducing auxiliary predictive information, rather than from any particular model architecture. SADQ only requires a one-step predictive signal, and the relative overhead can be further mitigated in ensemble-DQN settings or pre-collected data scenarios, where the auxiliary model can be trained offline or reused more efficiently.

The scope of SADQ is motivated by the continuing role of bootstrapped TD targets in value-based reinforcement learning. Our experiments focus on DQN-based agents because they expose the max-based action selection step most directly, and distributional variants preserve the same target-selection issue when return distributions are used for action evaluation. This focus does not make the problem narrow. Recent work still revisits the stability, efficiency, and statistical behavior of deep TD learning and Q-based methods, suggesting that target construction remains a relevant design point in modern RL~\citep{korkmaz2026counteractive, gallici2025simplifying, kastner2025categorical}. SADQ contributes to this line by addressing how bootstrapped targets can be made less sensitive to noisy value estimates during action aggregation.

\section{Conclusion}
In this work, we revisited TD target construction through the perspective of action aggregation under uncertainty and identified hard maximization over noisy Q-values as a structural source of instability in deep Q-learning. Rather than redesigning value estimators, we introduced SADQ, a lightweight successor-rollout aggregation mechanism that regularizes greedy target formation while preserving the standard bootstrap framework. Theoretically, we show that the resulting mixed Bellman operator is pointwise no more optimistic than the standard backup and recovers the optimal Bellman target in the ideal limit as model error approaches zero. Empirically, across classical control tasks, real-world scenarios, and Atari100K benchmarks, SADQ consistently improves training stability and final performance, and integrates seamlessly as a plug-in modification to both DQN and distributional variants. These results suggest that restructuring the aggregation step of the TD update itself offers a simple and principled path toward more robust Q-learning. 

\begin{acknowledgements}
The work of Xiaonan Zhang is partially supported by NSF under grants CCF-2312617, CNS-2431553, and IIS-2544108. We thank the anonymous reviewers for their helpful feedback.
\end{acknowledgements}

\bibliography{uai2026-template}

\onecolumn
\title{Revisiting TD Target Aggregation under Uncertainty in Q-Learning\\(Supplementary Material)}
\maketitle
\appendix
\section{Theoretical analysis}
\label{appx:theory_proof}
\paragraph{Lemma 1 (Local state extrapolation).}
Under the neural tangent kernel (NTK) regime~\citep{jacot2018neural}, for any in-sample state-action pair $(s,a) \in \mathcal{D}$ and in-neighborhood state-action pair $(s_{\mathcal{M}}, a)$ such that $\|s - s_{\mathcal{M}} \| \leq \epsilon$, the value difference of the deep Q function can be bounded as:
\begin{equation}\label{eq:lemma1_appx}
    \|Q_{\phi}(s'_{\mathcal M},a') - Q_{\phi}(s',a')\| \leq  C\left(\sqrt{\min\left(\|s'\oplus a'\|,\|s'_{\mathcal M}\oplus a'\|\right)}\sqrt{\epsilon}+ 2\epsilon \right),
\end{equation}
where $\oplus$ denotes the vector concatenation operation, and C is a finite constant.

\proof
The lemma follows directly from Theorem 1 or Lemma 4 in \cite{li2023when}. Please refer to \cite{li2023when} for detailed proofs. In addition, NTK remains one of the most influential theoretical frameworks for analyzing the generalization of deep neural networks~\citep{mao2026adaptive}.
\qed

\paragraph{Assumption 1 (Controlled model-induced perturbation).}
Let $\mathcal{M}_\theta(\cdot\mid s,a)$ be a stochastic dynamics model trained on the replay buffer support $\mathcal{D}$. Assume that for all $(s,a)\in\mathcal{D}$,
\begin{equation}
\label{eq:assum-appx}
\mathbb{E}\Big[\|r_{\mathcal{M}} + \gamma \max_{a'} Q'_\phi(s'_{\mathcal{M}},a')-\big(r + \gamma \max_{a'} Q'_\phi(s',a')\big)\|\,\Big|\, (s,a)\Big]\;\le\;\epsilon_r + \gamma \epsilon_s,
\end{equation}
where $(r,s') \sim \mathcal{T}(\cdot\mid s,a)$ denotes the true transition dynamic and 
$(r_{\mathcal{M}},s'_{\mathcal{M}}) \sim \mathcal{M}_\theta(\cdot\mid s,a)$ denotes the learned stochastic model. Moreover, $\epsilon_r$ bounds the reward prediction error of the learned model, and $\epsilon_s$ bounds the state-induced value perturbation characterized by \textbf{Lemma~1}, respectively.

\paragraph{Definition 1:~(Successor-state aggregation).}
The standard TD update relies on the greedy evaluation
\(
\max_{a'} Q'_\phi(s', a')
\)
at the observed next state $s'$. Here, we additionally employ a stochastic dynamics model 
$\mathcal{M}_\theta$ to predict the one-step outcomes of all next state-action pairs. For each $a' \in \mathcal{A}$, let
\(
(r'_{\mathcal M}, s''_{\mathcal M})\sim\mathcal{M}_\theta(\cdot \mid s', a').
\)
We define the model-based Bellman estimate
\begin{equation}
\label{eq:model_bellman}
\widetilde{Q'}_\phi(s', a')=r'_{\mathcal M}+\gamma \max_{a''} Q'_\phi(s''_{\mathcal M}, a'').
\end{equation}
Let
\(
\hat{a}'=\arg\max_{a' \in \mathcal{A}}\widetilde{Q'}_\phi(s', a'),
\)
then the resulting mixed TD target is defined as
\begin{equation}
\label{eq:mixed_target}
\hat{y}=\alpha \max_{a'} Q'_\phi(s', a')+(1-\alpha)Q'_\phi(s', \hat{a}'),
\end{equation}
where $\alpha \in [0,1]$ controls the trade-off between the standard greedy target and the model-guided action selection.

\paragraph{Remark 1:~~}Path-consistency methods such as PCZero construct a sliding window that contains historical states and MCTS-scouted nodes~\citep{zhao2023generalized}. They then minimize the value variance within that window. The goal is to enforce the principle that ``values on one optimal path should be identical''~\citep{zhao2022efficient}. \textit{Definition~1} realizes a one-step aggregation analogue: given the model-generated successor evaluations 
\(
\{\widetilde{Q}_\phi(s',a')\}_{a'\in\mathcal A},
\)
we select 
\(
\hat{a}'=\arg\max_{a'\in\mathcal A}\widetilde{Q}_\phi(s',a'),
\)
that is, the action whose model-based Bellman estimate is most aligned with the greedy backup. The mixed target in Eq.~(\ref{eq:mixed_target}) then interpolates between the standard TD greedy evaluation and this model-guided successor aggregation. In this sense, the aggregation step plays a role analogous to the low-variance window averaging in PCZero, but at the finest temporal granularity of a single transition.

\paragraph{Theorem 1 (Bootstrap bias reduction and ideal-limit target consistency).}
Let $\mathcal{T}_{\mathrm{std}}$ and $\mathcal{T}_{\mathrm{mix}}$ denote the standard Bellman operator and the mixed operator induced by Eq.~(\ref{eq:mixed_target_correct}), respectively. Then, the following properties hold:
\begin{enumerate}
    \item For any bounded action-value function $Q$, the mixed operator satisfies
    \[
    \mathcal{T}_{\mathrm{mix}} Q \;\le\; \mathcal{T}_{\mathrm{std}} Q
    \quad \text{pointwise}.
    \]
    \item In the ideal limiting case where the model-induced perturbation approach zero in \textbf{Assumption 1}, then
    \[
    \lim_{k \to \infty}
    \big\|
    \mathcal{T}_{\mathrm{mix}}^{(k)} Q^* - \mathcal{T}^* Q^*
    \big\|_\infty
    = 0.
    \]
\end{enumerate}

\proof
Define the following Bellman operators acting on bounded functions 
$Q:\mathcal{S}\times\mathcal{A}\to\mathbb{R}$:
\begin{align}
(\mathcal{T}_{\mathrm{std}} Q)(s,a)
&:=
r(s,a)
+
\gamma\, \mathbb{E}_{s'}\big[ \max_{a'} Q(s',a') \big], \\
(\mathcal{T}_{\mathrm{mix}} Q)(s,a)
&:=
r(s,a)
+
\gamma\, \mathbb{E}_{s'}\big[
\alpha \max_{a'} Q(s',a')
+
(1-\alpha) Q(s',\hat a'(s'))
\big],
\end{align}
where $\hat a'(s')$ is defined via successor-state aggregation 
(Definition~1), and $\mathcal{T}^*$ denotes the optimal Bellman operator.

\textbf{(I) Bias monotonicity.}

For any fixed $Q$ and any $s'$,
\[
Q(s',\hat a'(s')) \le \max_{a'} Q(s',a').
\]
Hence,
\[
(\mathcal{T}_{\mathrm{mix}} Q)(s,a)
-
(\mathcal{T}_{\mathrm{std}} Q)(s,a)
=
\gamma (1-\alpha)\,
\mathbb{E}_{s'}\big[
Q(s',\hat a'(s'))
-
\max_{a'} Q(s',a')
\big]
\le 0.
\]
Therefore,
\[
\mathcal{T}_{\mathrm{mix}} Q
\le
\mathcal{T}_{\mathrm{std}} Q
\quad
\text{pointwise}.
\]
This establishes that the mixed operator cannot increase the bootstrap bias relative to the greedy operator.

\textbf{(II) Ideal case consistency.}

By Definition~1, the model-based one-step score is
\[
\widetilde Q(s',a')
=
r'_{\mathcal M}
+
\gamma \max_{a''} Q(s''_{\mathcal M},a'').
\]
Assumption~1 implies that for all $(s',a')$ in the replay support,
\[
\big|
\widetilde Q(s',a')
-
\big(r' + \gamma \max_{a''} Q(s'',a'')\big)
\big|
\le
\epsilon_r + \gamma \epsilon_s.
\]

If
\[
\lim_{k\to\infty}
(\epsilon_r^{(k)}+\gamma \epsilon_s^{(k)})
=
0,
\]
then
\[
\sup_{s,a}
\big|
(\mathcal{T}_{\mathrm{mix}}^{(k)} Q)(s,a)
-
(\mathcal{T}_{\mathrm{std}} Q)(s,a)
\big|
\;\longrightarrow\;
0.
\]

In particular,
\[
\lim_{k\to\infty}
\|
\mathcal{T}_{\mathrm{mix}}^{(k)} Q^*
-
\mathcal{T}^* Q^*
\|_\infty
=
0.
\]
Since $Q^*$ is the unique fixed point of the optimal Bellman operator $\mathcal{T}^*$, the limiting result indicates that the mixed target recovers the optimal Bellman target in the ideal zero-error case.
\qed

\section{Scenarios of CityFlow and O-Cloud}
\label{appx:realworld}
In Appendix B, we illustrate the scenarios of CityFlow and O-Cloud that are used in our evaluations in detail, including the system model, state representation, action space, reward function, and setup. 
\subsection*{B.1. CityFlow Scenario}
We consider a traffic management scenario within the CityFlow simulation environment\footnote{https://github.com/cityflow-project/CityFlow/}, where a set of intersections, denoted as $\mathcal{C} = \{1, \cdots, c, \cdots, C\}$, are managed by a centralized traffic control system. Each intersection is equipped with traffic lights that regulate the flow of vehicles across various road networks. The goal is to manage traffic flow efficiently by adjusting the traffic light phases at each intersection based on real-time traffic conditions.

\textit{State Representation:\ \ }The state space, denoted as \(\mathcal{S}\), encapsulates crucial operational metrics:
\begin{itemize}
    \item Traffic Light Phases: Each intersection $c \in \mathcal{C}$ has multiple phases representing different traffic light states (e.g., green, yellow, red) that control the flow of vehicles. The state captures the current phase at each intersection.
    \item Vehicle Count on Lanes: The number of vehicles present on each lane leading into an intersection, represented by a vector \(V_{lane}\). This includes both the total number of vehicles and those waiting to pass through the intersection.
    \item Waiting Vehicle Count on Lanes: The number of vehicles waiting at each lane, represented by a vector \(W_{lane}\). This reflects the congestion level at the intersection.
\end{itemize}

\textit{Action Space:\ \ }The action space, \(\mathcal{A}\), involves selecting an appropriate traffic light phase for each intersection \(c \in \mathcal{C}\). The action taken at each step is to choose the phase that will be applied to control the traffic flow.

\textit{Reward Function:\ \ }The reward function is designed to minimize traffic congestion and vehicle wait times at intersections. The reward is computed based on the difference in the number of waiting vehicles before and after a traffic light phase change:

\begin{equation}
    r_{\text{congestion}}^c = - \sum_{l \in \mathcal{L}} \left(W_{lane}^{\text{after}} - W_{lane}^{\text{before}}\right),
\end{equation}

where \(\mathcal{L}\) represents the set of lanes at intersection \(c\). This reward structure encourages actions that reduce the number of waiting vehicles, thus alleviating congestion.

The overall reward for the environment at any given time is the sum of the rewards across all intersections:

\begin{equation}
    r_{\text{total}} = \sum_{c=1}^{C} r_{\text{congestion}}^c.
\end{equation}

\subsection*{B.2. O-Cloud Scenario}
We consider the computational task management on O-Cloud clusters, where a set of servers $\mathcal{M} = \{1, \cdots, m, \cdots, M\}$ with limited computing resources handle computing requests from applications (herein and after: users). The requests from different users are with various attributes in terms of CPU and RAM demands as well as the required processing latency. Upon the arrival of a user request at the O-Cloud, we assign it to an appropriate server for execution. In instances where a server is operating at full load to process concurrent requests, incoming requests are queued for temporary storage.

This scenario leverages the Alibaba cluster-trace-v2018 dataset~\footnote{https://github.com/alibaba/clusterdata}, an open-source collection of real production cluster workload traces. Spanning an 8-day period and encompassing data from 4000 machines, this dataset provides a detailed view of server characteristics, including CPU, memory, and communication bandwidth. Each task trace records arrival time, duration, and CPU resource demand, with tasks arranged in chronological order of arrival.

\textit{State Representation:\ \ }The state space, denoted as \(\mathcal{S}\), encapsulates vital operational metrics:
\begin{itemize}
    \item Demands of the incoming task (user request), including requirement of CPU (\(c_{\text{req}}\)) and RAM (\(r_{\text{req}}\)), as well as the estimated occupation time (\(t_{\text{occ}}\)).
    \item The CPU and RAM utilization rates for each server, represented as the vector \({U}_{\text{cpu}} = \{u_{\text{cpu}}^m|m\in \mathcal{M}\}\) and \({U}_{\text{ram}} = \{u_{\text{ram}}^m|m\in \mathcal{M}\}\), respectively. \({U}_{\text{cpu}}\) and \({U}_{\text{ram}}\)  indicate the current resource load.
    \item The length of the pending queue in each server, represented as the vector \(L_\text{queue} = \{l_{\text{queue}}^m|m\in \mathcal{M}\}\), reflecting the count of pending tasks when CPU and RAM are used to handle other tasks in server $m \in \mathcal{M}$.
    \item A dynamically calculated queue penalty vector \(P_{\text{queue}} = \{p_{\text{queue}}^m|m\in \mathcal{M}\}\) from each server.  \(P_{\text{queue}}\) quantifies the delay-induced penalty associated with tasks in the pending queue, where $p_{\text{queue}}^m = \sum\nolimits_{l=1}^{l_{\text{queue}}^m} t_{\text{occ}}^l$.   
\end{itemize}
The queue penalty, \(P_{\text{queue}}\), for server $m$ is calculated based on the resource demands and occupation time of tasks in each server's pending queue. Specifically,
\begin{equation}
    P_{\text{queue}}^m = \sum_i (c_{\text{req}}^i + r_{\text{req}}^i) \cdot t_{\text{occ}}^i.
\end{equation}
This penalty measure helps in understanding the resource demand and processing backlog of tasks queued for execution, aiding in making more informed server selection decisions.

\textit{Action Space:\ \ } The action space, \(\mathcal{A}\), involves selecting an appropriate server for each incoming task, which can be mathematically represented as choosing a server \(a \in \{1, 2, ..., M\}\) for task allocation.

\textit{Reward Function:\ \ }The design of the reward function seeks to concurrently minimize power consumption and reduce the latency for users. For a set of \( M \) servers, instantaneous power \( P_{\text{power}} \) is calculated based on the CPU utilization rates of the individual servers. This is given by~\citep{liu2017hierarchical} and we modify it as:
\begin{equation}
    P_{\text{power}} = \sum_{m=1}^{M} \left(P_0 + (P_{1} - P_0)*( 2u_{\text{cpu}}^m - (u_{\text{cpu}}^m)^{1.4}) / P_{1} \right)
\end{equation}
Suppose that the server $m$ is selected to handle the incoming task, the delay penalty \( P_{\text{latency}} \) is defined as a normalized measure based on the queuing penalties across all servers:
\begin{equation}
P^{cur}_{\text{latency}} = \sum_{i=0}^{cur_{time}} 
\begin{cases} 
t^i_{latency}, & if\ task\ starts \\
cur - t^i_{arr}, & else
\end{cases}
\end{equation}

where \(cur\) denotes the current time, \(t^i_{latency}\) is the latency of task \(i\), and \(t^i_{arr}\) is the arrival time of task \(i\). Then, we can compute the reward for latency at current time:
\begin{equation}
r_{latency} = P^{cur}_{\text{latency}} - P^{cur-1}_{\text{latency}}
\end{equation}

The reward function \( r_x \) is then expressed as a weighted sum of the instantaneous power among all servers and the current latency penalty to a selected server, defined as:
\begin{equation}\label{Eq:orireward}
    r_x = -\left(w_1 \cdot P_{\text{power}} + w_2 \cdot r_{latency}\right),
\end{equation}
where \(w_1\) and \(w_2\) are weighting coefficients. 

\section{Experimental Configurations}
\label{appx:experiments_config}
All vector-based RL environments and baselines are implemented in DI-Engine\footnote{\url{https://github.com/opendilab/DI-engine}}. 
All image-based RL environments and baselines are implemented in Jax-baseline~\footnote{\url{https://github.com/tinker495/jax-baseline}}.
All experiments are conducted on a machine equipped with an NVIDIA RTX 5090 GPU, an Intel Core i9-14900k processor, and 128 GB of DDR5 RAM.

\noindent\textbf{Vector-based Tasks:}
Table~\ref{tab:combined_env_config} provides detailed configurations for both Q and \(\mathcal{M}\) on vector-based tasks. The term \textit{Update per Collect} refers to the number of training steps performed after every \textit{Replay Frequency} steps. The term \textit{Target Update Interval} indicates the frequency, in steps, at which the target Q-network is updated. Parameters under \textit{Epsilon} correspond to the settings of the epsilon-greedy exploration strategy.
\begin{table*}[ht]
    \centering
    \caption{Training Configurations for Q and \(\mathcal{M}\) on vector-based tasks.}
    \label{tab:combined_env_config}
    \begin{tabular}{lcccccc}
        \toprule
        \textbf{Parameter} & \textbf{Acrobot-V1} & \textbf{LunarLander-V2} & \textbf{Cartpole-V0} & \textbf{BitFlip} & \textbf{O-Cloud} & \textbf{CityFlow} \\
        \midrule
        \multicolumn{7}{c}{\textbf{Configurations for Q}} \\
        \midrule
        Discount               & 0.99       & 0.99       & 0.97           & 0.99           & 0.8      & 0.99       \\
        Hidden Size            & [256, 256] & [512, 64]  & [128, 128, 64] & [128, 128, 64] & [64, 64] & [256, 256] \\
        Batch Size             & 128        & 64         & 64             & 128            & 32       & 64         \\
        Learning Rate          & 1e-4       & 1e-3       & 1e-3           & 5e-4           & 5e-5     & 5e-5       \\
        Update per Collect     & 10         & 10         & 1              & 10             & 1        & 1          \\         
        Target Update Interval & 2400       & 640        & 8000           & 4800           & 2000     & 2000       \\
        Total Steps            & 960000     & 128000     & 160000         & 960000         & 500000   & 400000     \\
        Buffer Size            & 100000     & 100000     & 100000         & 4000           & 100000   & 100000     \\
        Replay Frequency       & 96         & 64         & 80             & 96             & 100      & 100        \\
        Epsilon Start          & 1          & 0.95       & 0.95           & 0.2            & 0.05     & 0.05       \\         
        Epsilon End            & 0.05       & 0.1        & 0.1            & 0.2            & 0.05     & 0.05       \\         
        Epsilon Decay          & 250000     & 50000      & 10000          & 100            & 10000    & 10000      \\         
        \midrule
        \multicolumn{7}{c}{\textbf{Configurations for \(\mathcal{M}\)}} \\
        \midrule
        Hidden Size            & [256, 256] & [256, 256] & [256, 256]     & [256, 256]     & [64, 64] & [256, 256] \\
        Batch Size             & 256        & 128        & 128            & 256            & 64       & 128        \\
        Learning Rate          & 4e-5       & 4e-5       & 4e-5           & 4e-4           & 5e-4     & 5e-4       \\
        Update per Collect     & 1          & 1          & [1,5,10,20]    & 1              & 1        & 1          \\         
        State Norm             & 1          & 1          & 1              & 1              & 50       & 15         \\
        Factor \(\alpha\)      & 0.2        & 0.2        & 0.2            & 0.2            & 0.5      & 0.5        \\      
        \bottomrule
    \end{tabular}
\end{table*}

For the BitFlip environment, we set \(n_{\text{bits}} = 8\). The cloud environment comprises 10 servers, resulting in an action space size of \(|\mathcal{A}| = 10\). The reward function is parameterized by weights \(w_1 = 0.1\) and \(w_2 = 0.005\). The system begins with a warm-up phase of 1000 tasks, followed by 200 user requests. In the CityFlow simulation environment, there are 4 intersections, each with 4 control phases, yielding an action space of size \(|\mathcal{A}| = 256\). The simulation operates over a fixed episode duration, where each step represents a predefined time period during which traffic light phases can be adjusted.

\noindent\textbf{Image-based Tasks (Atari100K):}
For Atari100K experiments, we follow a standard value-based training setup. We use AdamW optimizer with a learning rate of $1\times10^{-4}$. The Q-network consists of a 2048-dimensional hidden layer. Training is performed for 100k environment steps with batch size 32 and training frequency 4. We use $\gamma=0.99$, replay buffer size $10^5$, and start learning after 1,600 steps. Exploration is annealed over 40\% of training to a final $\epsilon$ of 0.001. The target network is updated every 1,000 steps, and two gradient updates are performed per environment interaction.

For the one-step RSSM-style dynamics model, the latent dynamics consists of a deterministic state of dimension 4096 and a categorical stochastic state with $32$ variables and $32$ classes per variable. The model uses a hidden size of 512, with AdamW using learning rate $3\times10^{-4}$ and weight decay $10^{-4}$. The training objective combines dynamics, representation, reconstruction, and reward terms with weights $\lambda_{\mathrm{dyn}}=1.0$, $\lambda_{\mathrm{rep}}=0.1$, $\lambda_{\mathrm{recon}}=2.0$, and $\lambda_{\mathrm{rew}}=1.0$, and uses free nats of $1.0$. For the adaptive dynamics-model update schedule, we set the reconstruction-loss threshold to $\tau_{\mathcal{M}}=0.001$ and the reduced relative update frequency to $t_f=0.4$. For SADQ, we set the mixed-target coefficient to $\alpha=0.2$ in all image-based experiments.

\section{Controlled Baseline Comparisons}
\label{appx:baseline_comparison}

\begin{figure}[H]
\centering
\subfloat[Acrobot]{
    \includegraphics[width=0.275\textwidth]{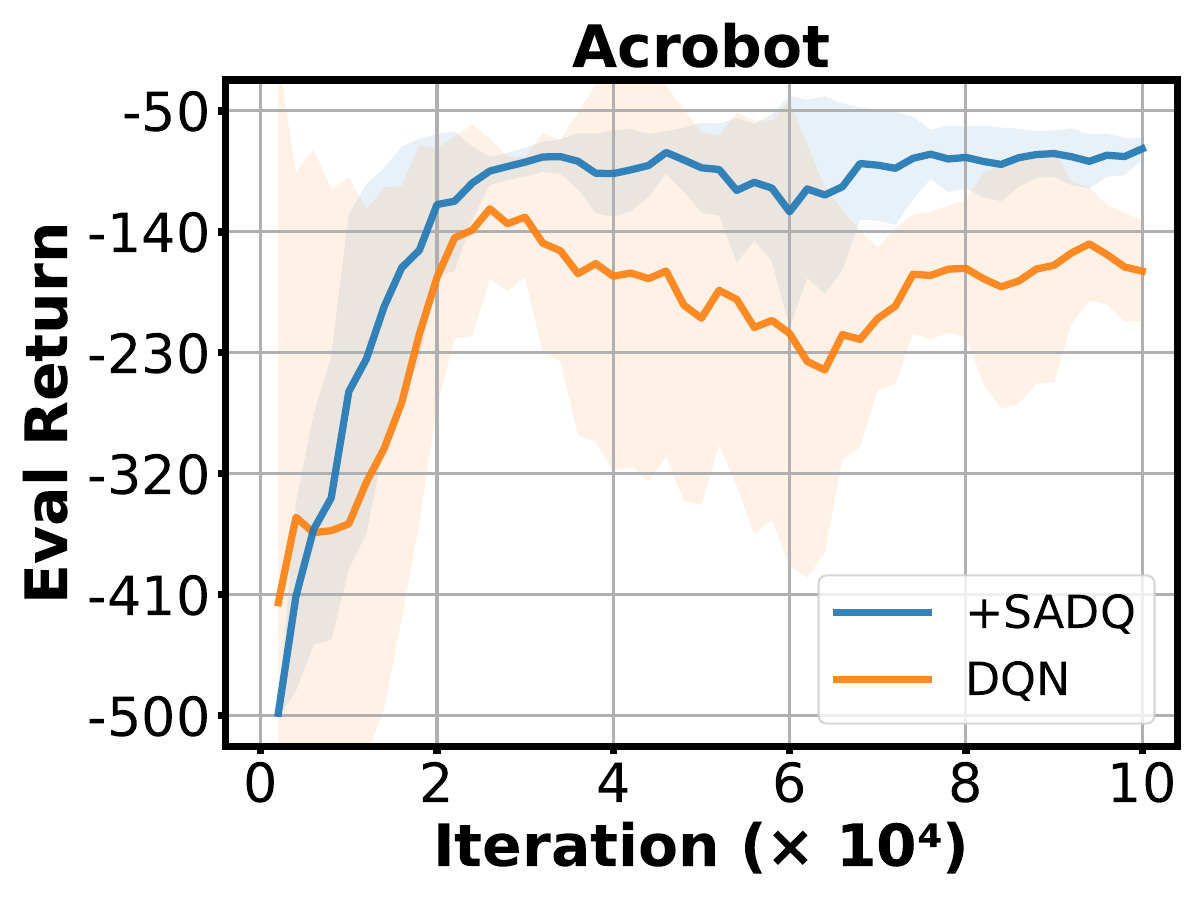}
}
\hfill
\subfloat[BitFlip]{
    \includegraphics[width=0.275\textwidth]{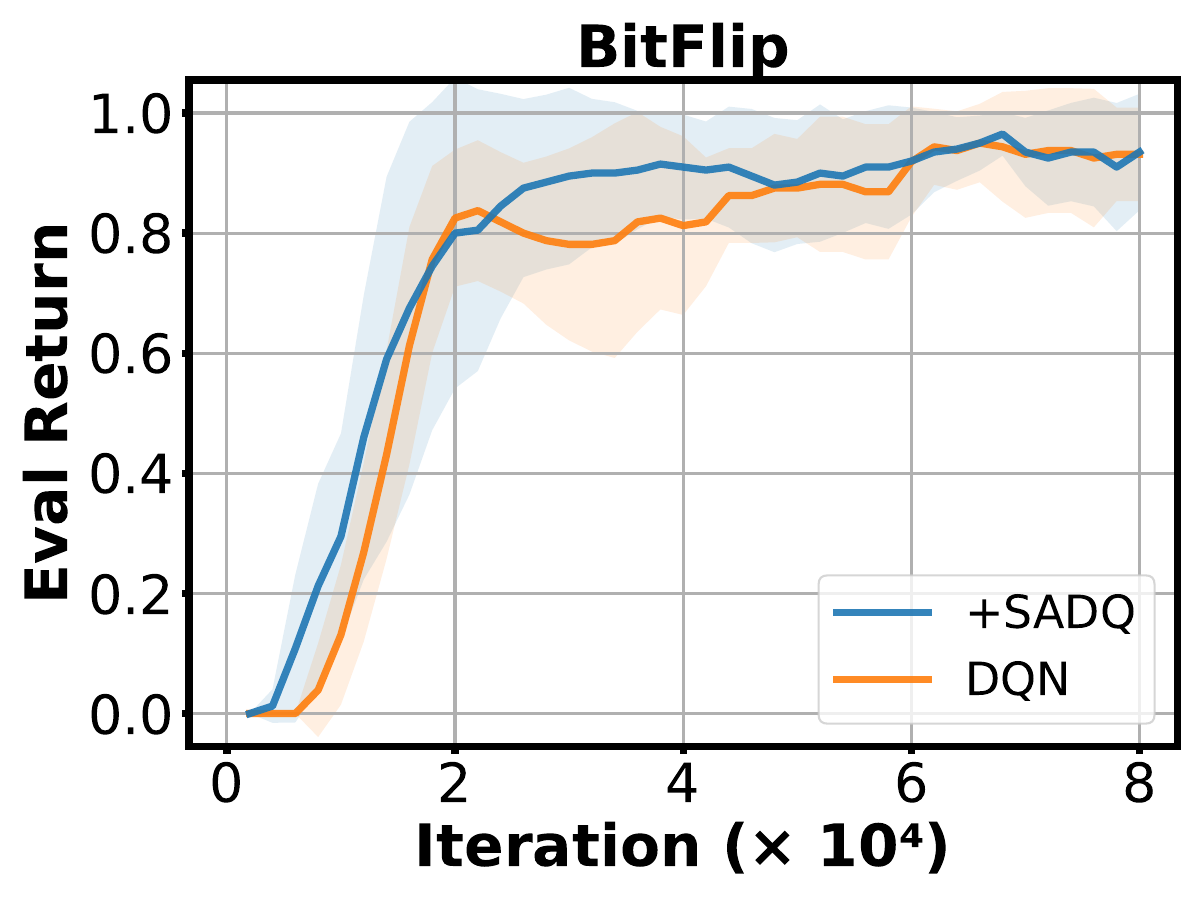}
}
\hfill
\subfloat[LunarLander]{
    \includegraphics[width=0.275\textwidth]{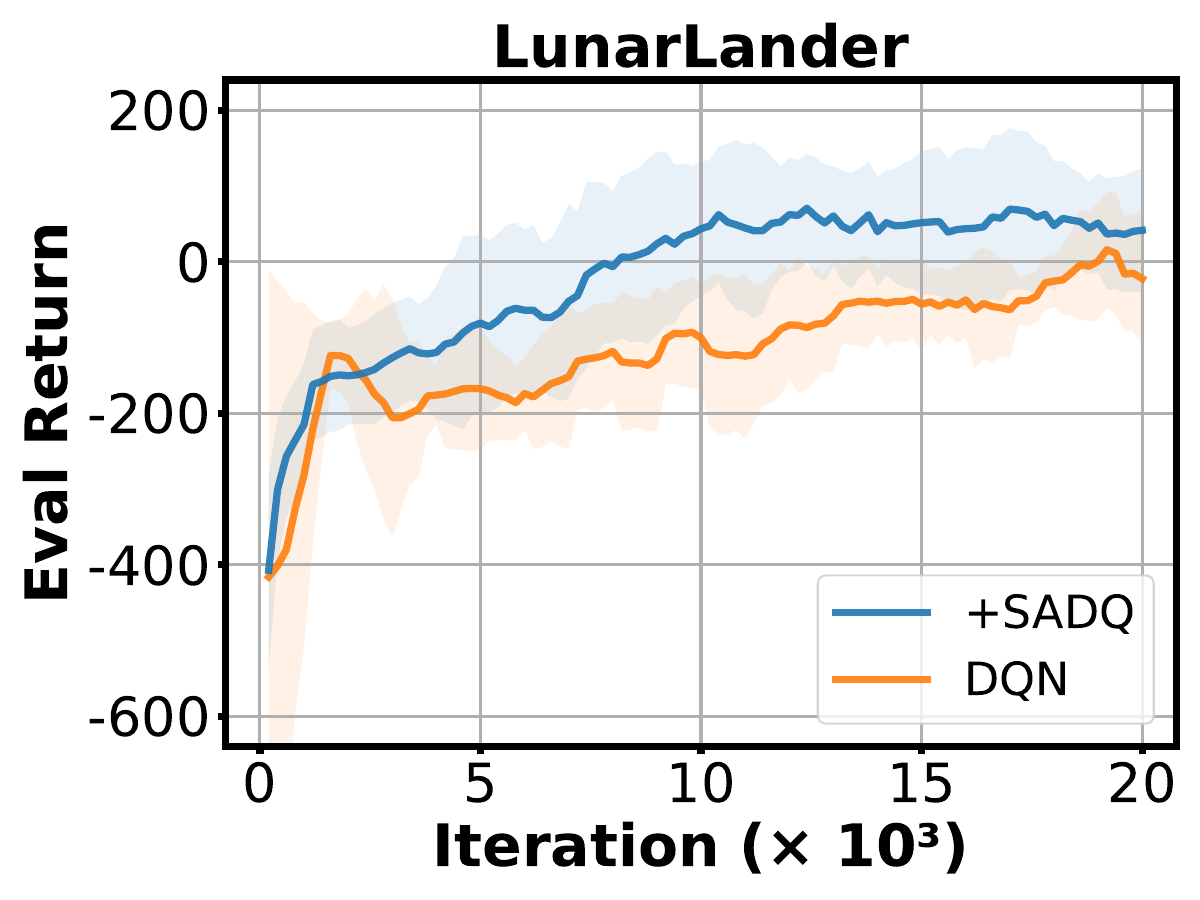}
}
\caption{Controlled comparison of DQN and DQN+SADQ across Acrobot, BitFlip, and LunarLander.}
\label{fig:dqn_sadq_comparison}
\end{figure}

\begin{figure}[H]
\centering
\subfloat[Acrobot]{
    \includegraphics[width=0.275\textwidth]{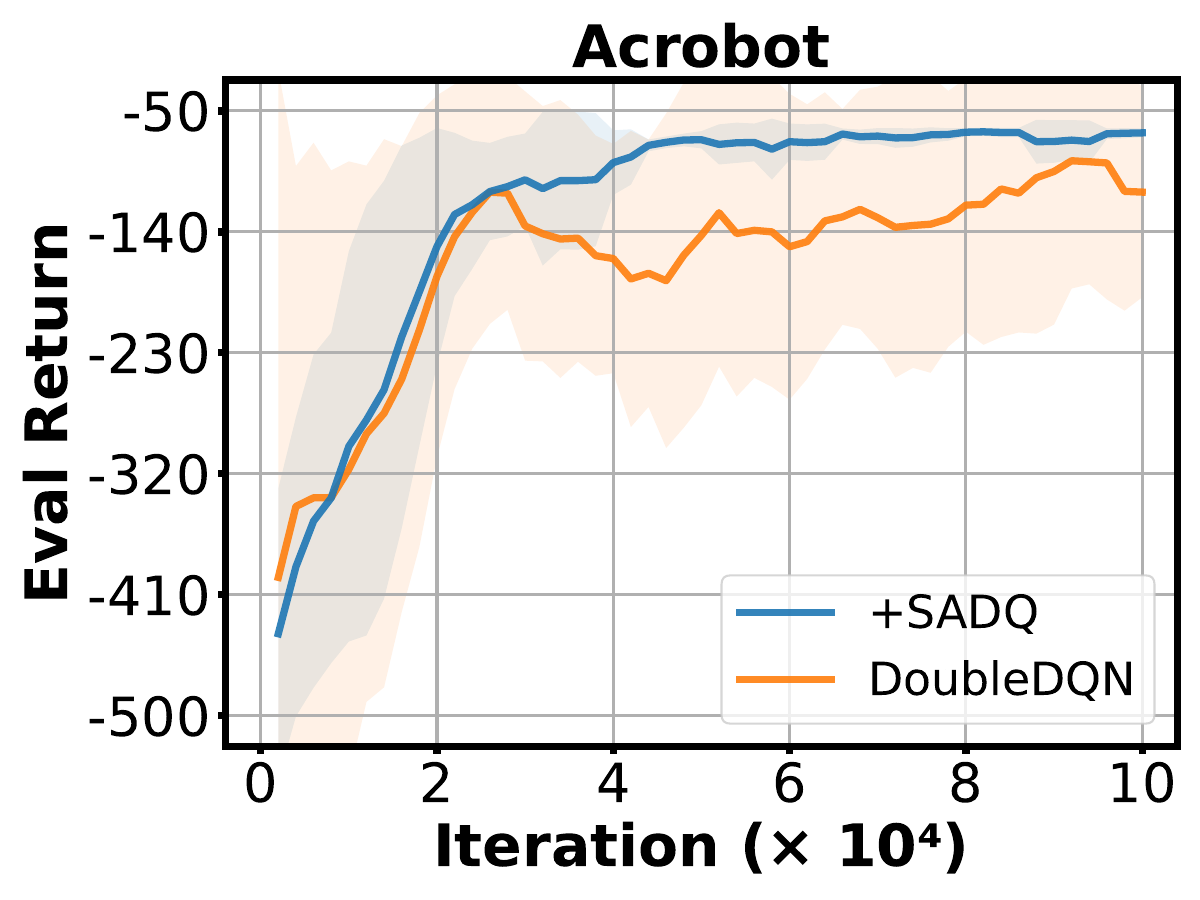}
}
\hfill
\subfloat[BitFlip]{
    \includegraphics[width=0.275\textwidth]{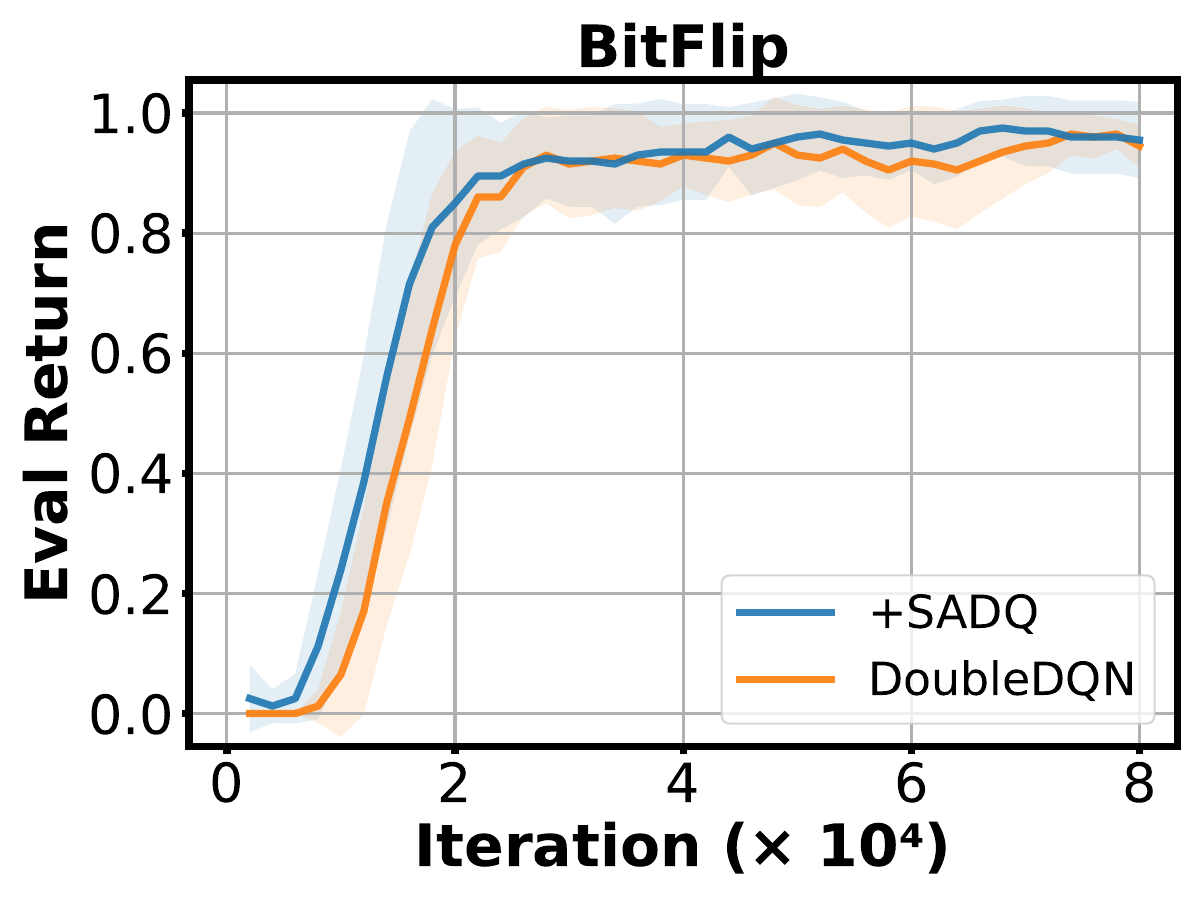}
}
\hfill
\subfloat[LunarLander]{
    \includegraphics[width=0.275\textwidth]{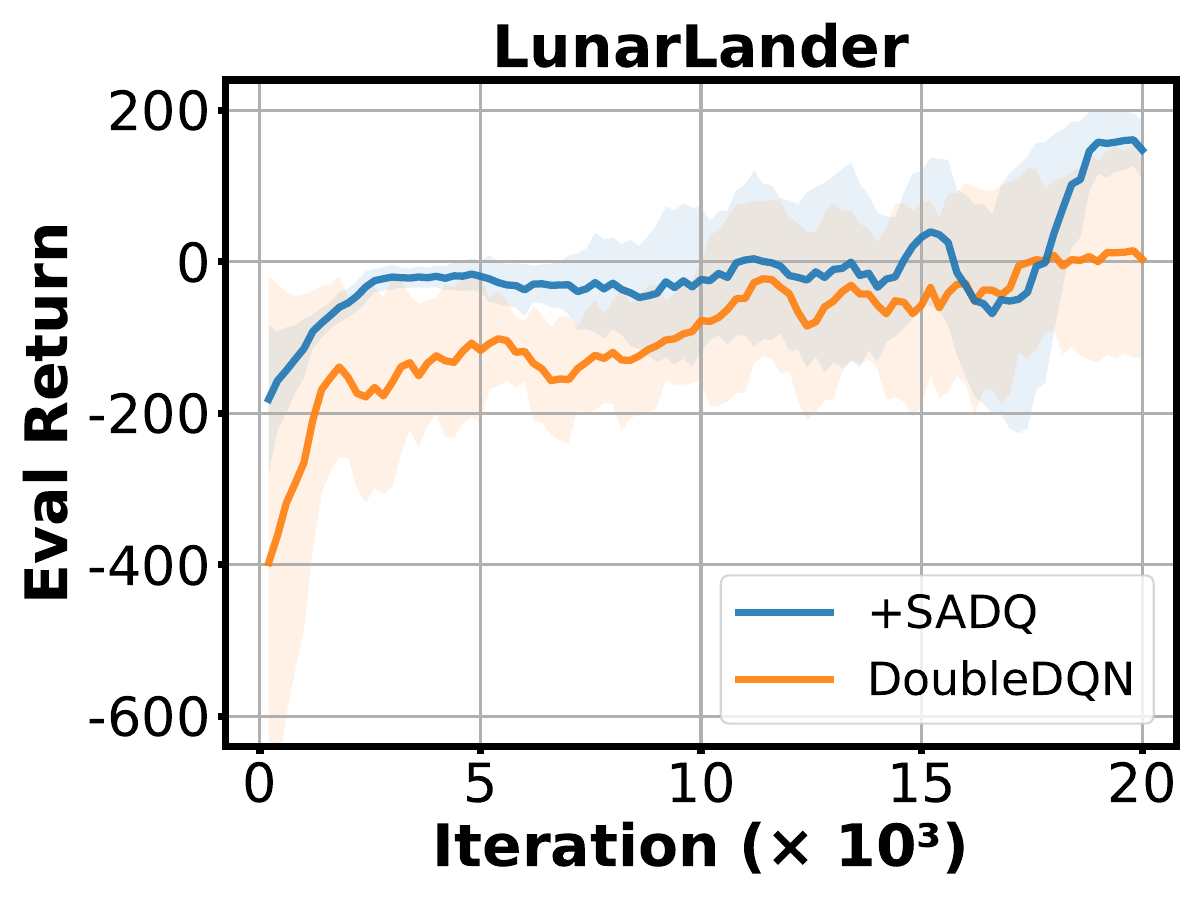}
}
\caption{Controlled comparison of Double DQN and Double DQN+SADQ across Acrobot, BitFlip, and LunarLander.}
\label{fig:double_dqn_sadq_comparison}
\end{figure}

\begin{figure}[H]
\centering
\subfloat[Acrobot]{
    \includegraphics[width=0.275\textwidth]{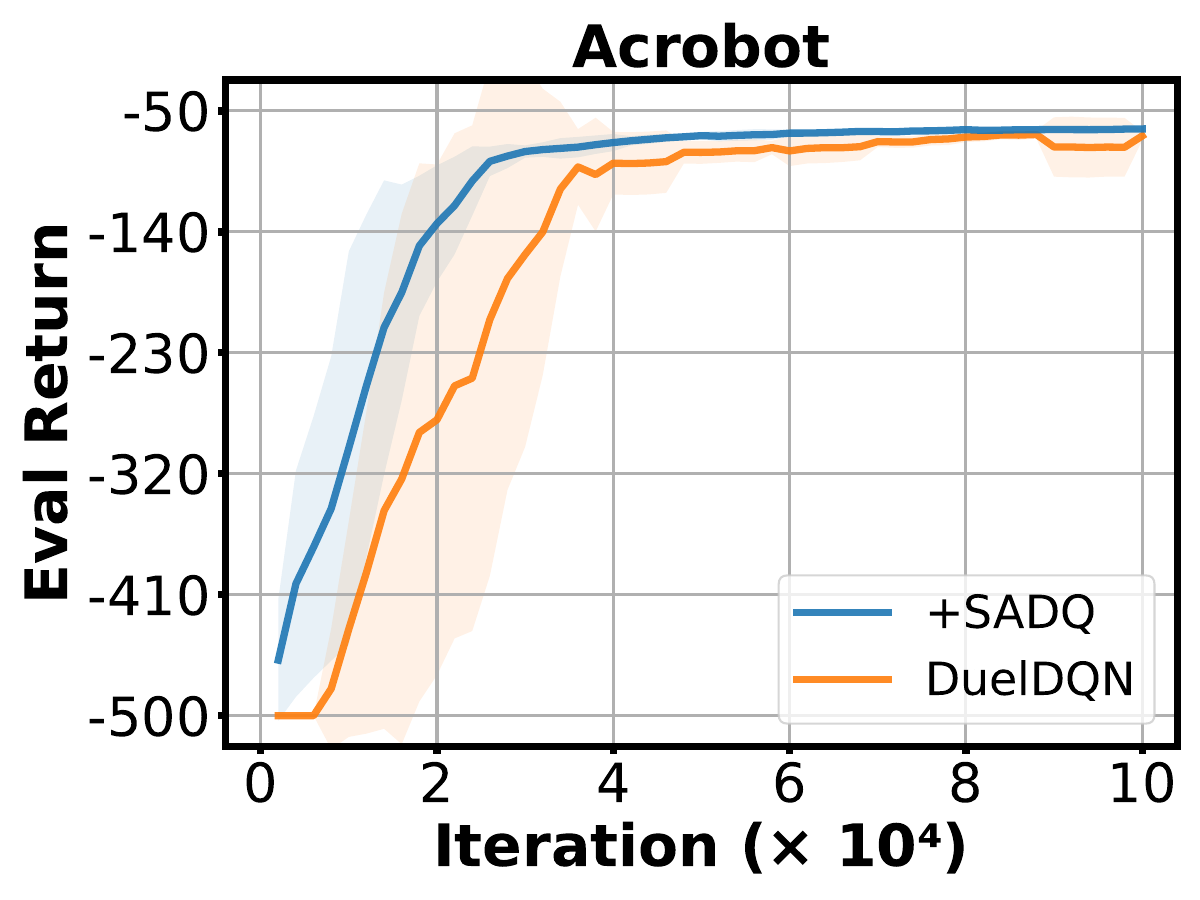}
}
\hfill
\subfloat[BitFlip]{
    \includegraphics[width=0.275\textwidth]{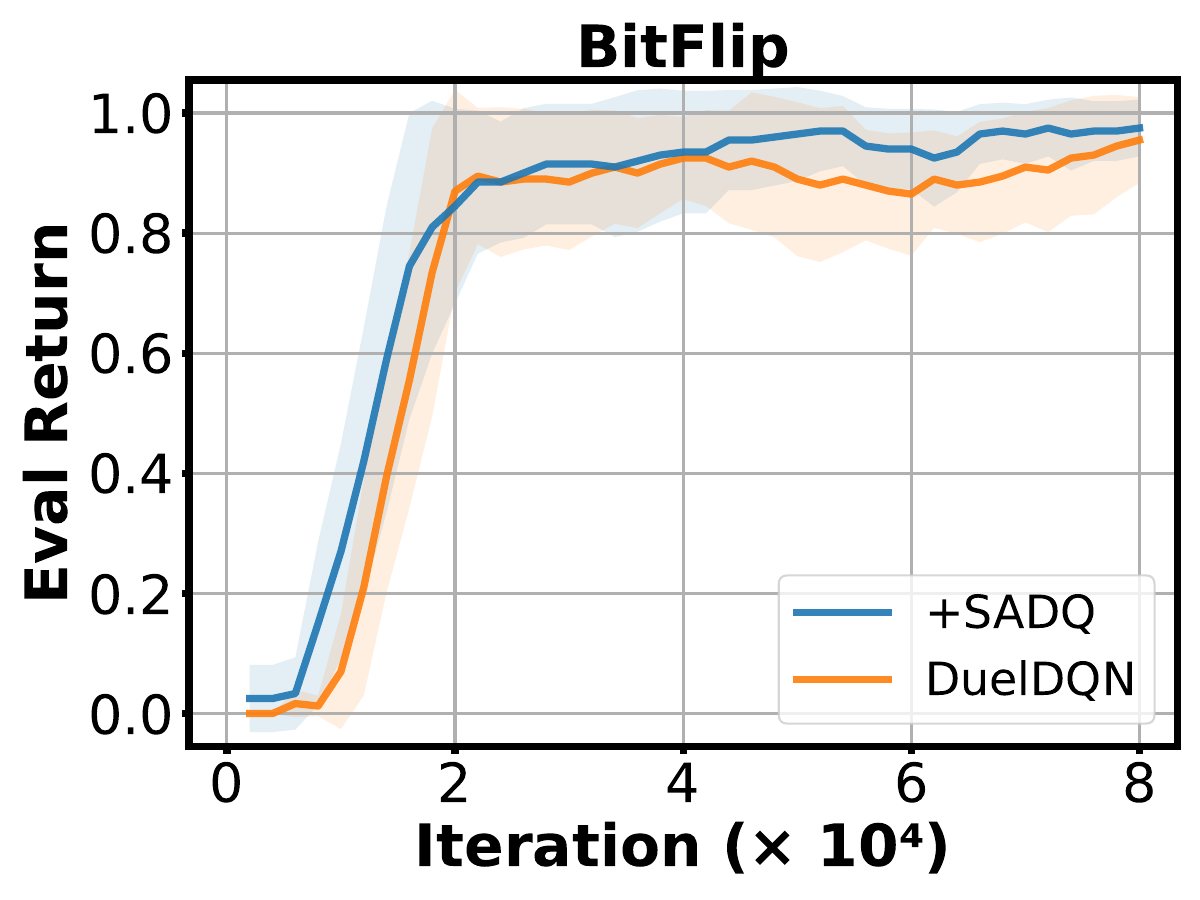}
}
\hfill
\subfloat[LunarLander]{
    \includegraphics[width=0.275\textwidth]{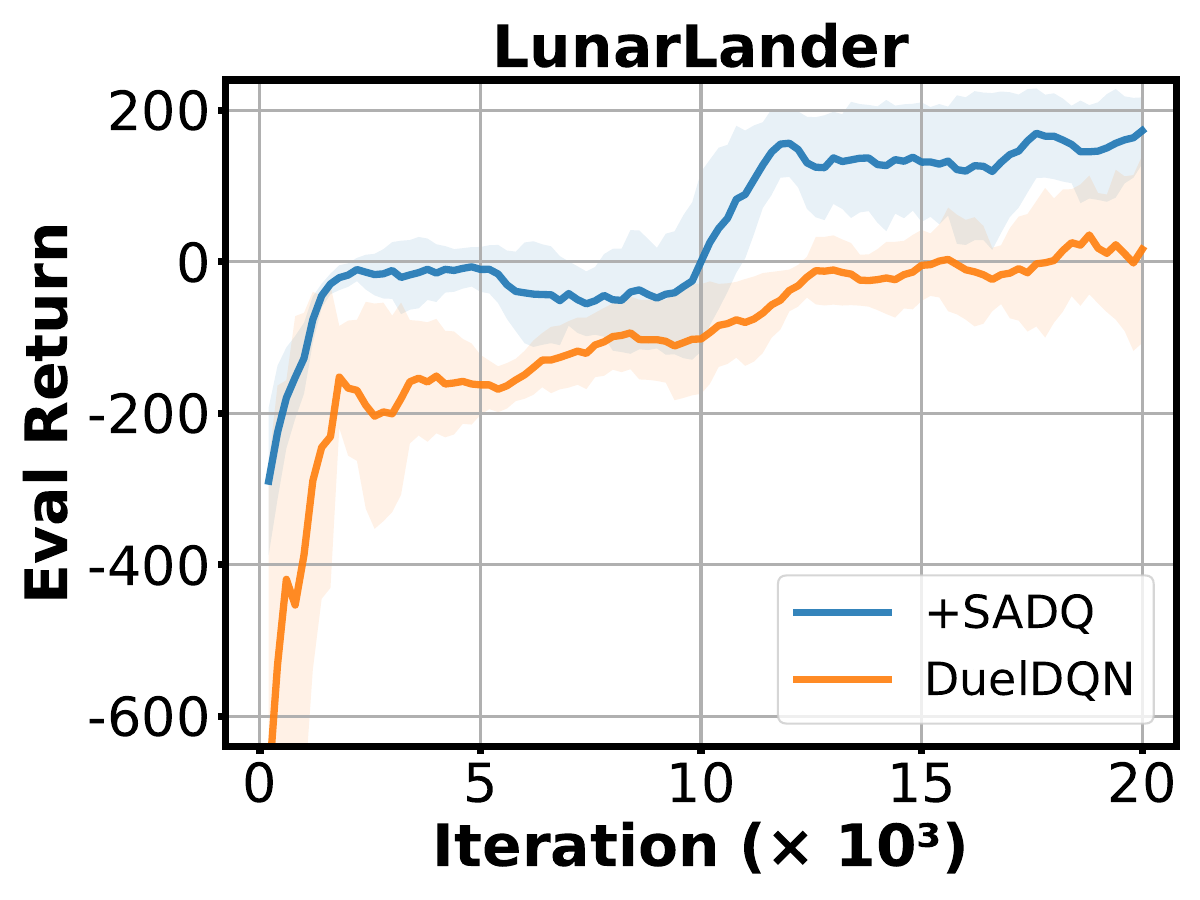}
}
\caption{Controlled comparison of Dueling DQN and Dueling DQN+SADQ across Acrobot, BitFlip, and LunarLander.}
\label{fig:dueling_dqn_sadq_comparison}
\end{figure}

\begin{figure}[H]
\centering
\subfloat[Acrobot]{
    \includegraphics[width=0.275\textwidth]{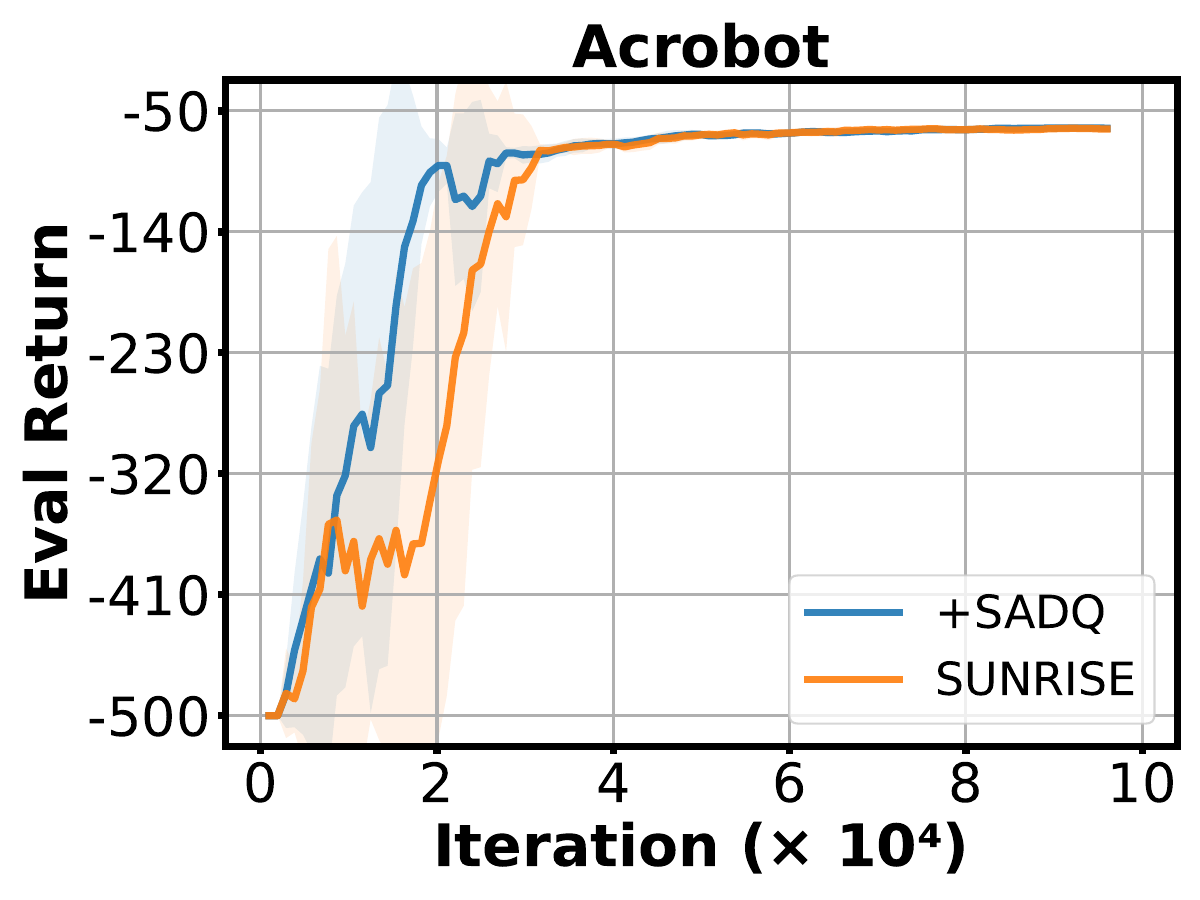}
}
\hfill
\subfloat[BitFlip]{
    \includegraphics[width=0.275\textwidth]{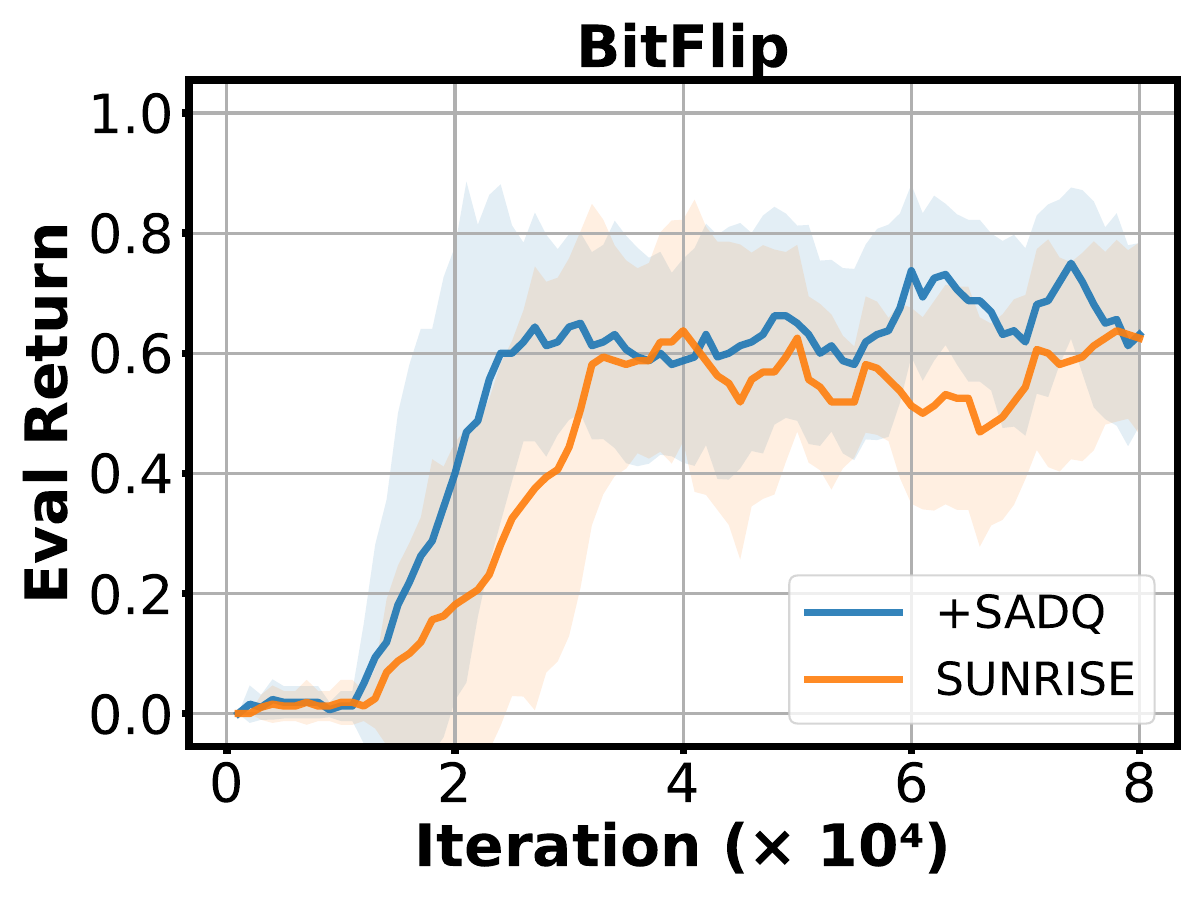}
}
\hfill
\subfloat[LunarLander]{
    \includegraphics[width=0.275\textwidth]{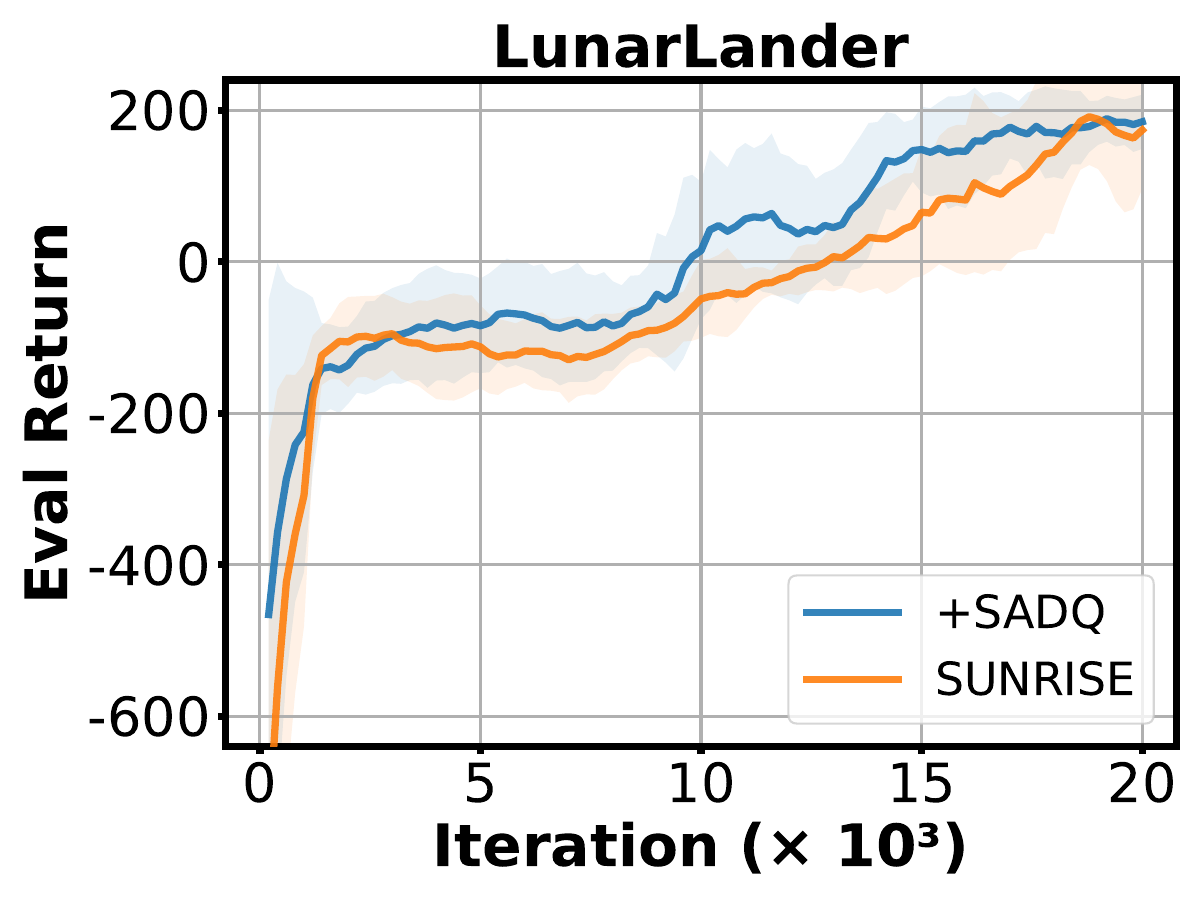}
}
\caption{Controlled comparison of SUNRISE and SUNRISE+SADQ across Acrobot, BitFlip, and LunarLander. SUNRISE uses an ensemble size of 5.}
\label{fig:sunrise_sadq_comparison}
\end{figure}

To isolate the contribution of SADQ, we compare multiple baselines with and without SADQ under identical experimental settings. The only difference is the max-operator-related component introduced by SADQ. Figs~\ref{fig:dqn_sadq_comparison}--\ref{fig:sunrise_sadq_comparison}
show the controlled comparisons between each baseline and its SADQ variant. SADQ consistently improves all baselines, suggesting that the proposed component is not tied to a specific DQN variant.

\section{Performance Comparison of SADQ with Other Baselines}
\label{appx:cartpole}
\begin{figure}[H]
    \centering
    \includegraphics[width=0.28\textwidth]{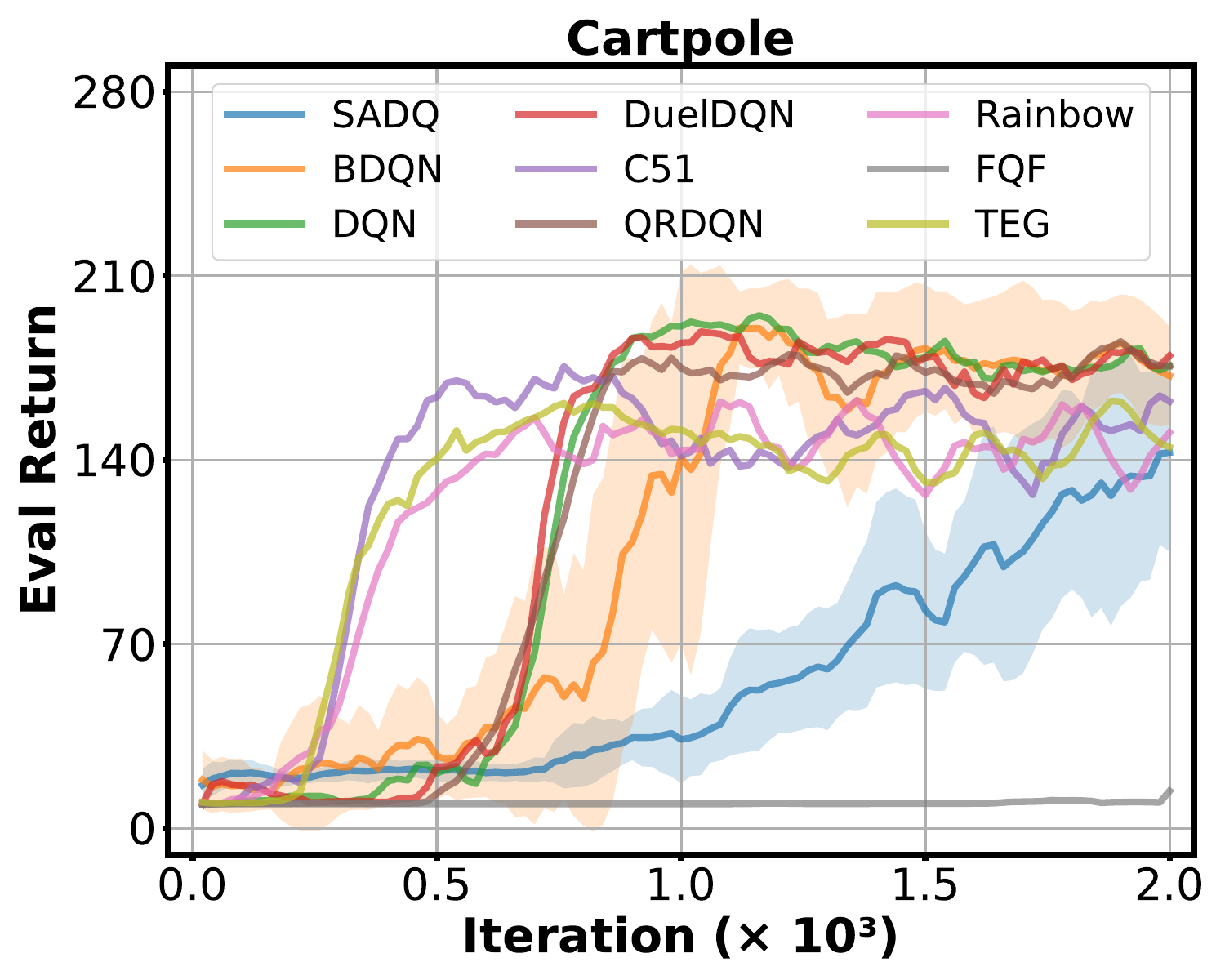}
    \caption{Performance comparison of SADQ with other baselines in Cartpole.}
    \label{fig:cartpole_appendix}
\end{figure}

As discussed in the main text, SADQ is shown to significantly underperform compared to all baseline algorithms in the Cartpole environment in Fig.~\ref{fig:cartpole_appendix}. While baselines like DQN and others achieve acceptable results, SADQ struggles to learn an effective policy and fails to match even the simplest algorithms. The simplicity of the Cartpole environment explains this outcome: with only two discrete actions and a state space of four dimensions, Cartpole poses minimal complexity for learning. Most baseline methods converge to a satisfactory policy within a few hundred iterations. Even advanced DQN variants, despite potentially introducing noise due to their complexity, are able to learn an acceptable policy given the short training time required by Cartpole.

On the other hand, SADQ’s reliance on the stochastic model \(\mathcal{M}\) for successor state predictions introduces a critical limitation. For \(\mathcal{M}\) to make accurate predictions, it requires sufficient training time and data to stabilize. In environments with shorter learning horizons like Cartpole, \(\mathcal{M}\) does not have enough iterations to converge effectively, leading to poor performance. This dependency on the stochastic model highlights a trade-off in SADQ's design: while it excels in tasks requiring detailed environment dynamics modeling, it may struggle in tasks where learning needs to occur rapidly. This observation sheds light on a key aspect of SADQ.  

\end{document}